\documentclass[journal]{IEEEtran}
\makeatletter
\long\def\@makecaption#1#2{\ifx\@captype\@IEEEtablestring
\footnotesize\begin{center}{\normalfont\footnotesize #1}\\
{\normalfont\footnotesize\scshape #2}\end{center}%
\@IEEEtablecaptionsepspace \else \@IEEEfigurecaptionsepspace
\setbox\@tempboxa\hbox{\normalfont\footnotesize {#1.}~~ #2}
\ifdim \wd\@tempboxa >\hsize%
\setbox\@tempboxa\hbox{\normalfont\footnotesize {#1.}~~ }
\parbox[t]{\hsize}{\normalfont\footnotesize \noindent\unhbox\@tempboxa#2}
\else \hbox
to\hsize{\normalfont\footnotesize\hfil\box\@tempboxa\hfil}\fi\fi}
\makeatother
\usepackage{array}
\usepackage{booktabs}
\usepackage{multirow}
\usepackage{amsmath}
\usepackage{amssymb}
\usepackage{amsthm}
\usepackage{graphicx}
\usepackage{cite}
\usepackage{subfigure}
\usepackage{url} 
\usepackage[top=2cm, bottom=2cm, left=2cm, right=2cm]{geometry}
\usepackage{algorithm}
\usepackage{algorithmic}
\usepackage{url}

\theoremstyle{definition} \newtheorem{theorem}{Theorem}
\newtheorem{definition}{Definition} 

\DeclareMathOperator*{\argmin}{\arg\min}

\ifCLASSINFOpdf

\else

\fi

\usepackage{bm}
\usepackage{amsfonts}

\begin{document}
%
\title{Fast Disentangled Slim Tensor Learning for Multi-view Clustering}
%
%
%

\author{
	Deng Xu,
	Chao Zhang, 
	Zechao Li,
	Chunlin Chen, 
	and Huaxiong Li
\thanks{
This work was supported in part by the National Natural Science Foundation of China under Grants 62176116, 62276136, 62073160, and in part by the National Key Research and Development Program of China under Grant 2022ZD0118802. \emph{(Corresponding author: Chao Zhang.)}

Deng Xu, Chao Zhang, Chunlin Chen, and Huaxiong Li are with the Department of Control Science and Intelligence Engineering, Nanjing University, Nanjing 210093, China (e-mail: dengxu@smail.nju.edu.cn, chzhang@smail.nju.edu.cn, clchen@nju.edu.cn, huaxiongli@nju.edu.cn).

Zechao Li is with the School of Computer Science and Engineering, Nanjing
University of Science and Technology, Nanjing 210014, China. (e-mail:
zechao.li@njust.edu.cn).
}}
%
%

\markboth{IEEE TRANSACTIONS ON MULTIMEDIA}%
{Shell \MakeLowercase{\textit{et al.}}: Bare Demo of IEEEtran.cls
for Journals}
%



\maketitle

\begin{abstract}
Tensor-based multi-view clustering has recently received significant attention due to its exceptional ability to explore cross-view high-order correlations. However, most existing methods still encounter some limitations. (1) Most of them explore the correlations among different affinity matrices, making them unscalable to large-scale data. (2) Although some methods address it by introducing bipartite graphs, they may result in sub-optimal solutions caused by an unstable anchor selection process. (3) They generally ignore the negative impact of latent semantic-unrelated information in each view. To tackle these issues, we propose a new approach termed fast Disentangled Slim Tensor Learning (DSTL) for multi-view clustering . Instead of focusing on the multi-view graph structures, DSTL directly explores the high-order correlations among multi-view latent semantic representations based on matrix factorization. To alleviate the negative influence of feature redundancy, inspired by robust PCA, DSTL disentangles the latent low-dimensional representation into a semantic-unrelated part and a semantic-related part for each view. Subsequently, two slim tensors are constructed with tensor-based regularization. To further enhance the quality of feature disentanglement, the semantic-related representations are aligned across views through a consensus alignment indicator. Our proposed model is computationally efficient and can be solved effectively. Extensive experiments demonstrate the superiority and efficiency of DSTL over state-of-the-art approaches. The code of DSTL is available at \url{https://github.com/dengxu-nju/DSTL}.
\end{abstract}

%
\IEEEpeerreviewmaketitle

\section{Introduction}\label{Sec1}

\IEEEPARstart{I}{n} many real-world applications, data can originate from different sources and feature collectors. For instance, we can utilize text posts, images, videos, user profiles, and social network connections to depict user behavior and interactions on social media platforms. To analyze sensor signals, we can decompose them into time and frequency domains. These are known as multi-view data, which often detail an object from various perspectives and provides a richer and more comprehensive understanding compared to single-view data. In recent years, the surge in multi-view data has sparked considerable interest in multi-view learning (MVL). MVL seeks to leverage the potential consistent and complementary information across multiple views to enhance generalization performance in downstream machine learning activities such as classification and clustering~\cite{10172029,pu2024federated,ji2023high,10506642,zhang2023enhanced,zhang2024learning,10232925,WenYZXWFZ21}. Single-View Clustering (SVC) refers to the clustering of data comprising of a single view, while Multi-View Clustering (MVC) is to partition the multi-view data, leading to superior results than those obtained with SVC~\cite{xia2022tensorized,chen2023fast,LiuWWXH23}.

Most existing methods for MVC have demonstrated success, such as those employing matrix factorization~\cite{zong2018multi,ma2021discriminative,liu2021one}, graph learning~\cite{TangZLLWZW19,huang2021measuring,wang2021fast}, and subspace learning~\cite{zhang2018generalized,sun2021scalable,chen2022efficient}. Multi-view matrix factorization (MultiMF) effectively reduces the dimensionality of high-dimensional data and captures diverse underlying representations of multiple views. For example, Ma et al.~\cite{ma2021discriminative} integrated multi-view linear discriminant analysis with MultiMF to leverage the intrinsic low-dimensional structure within the projection subspace. Liu et al.~\cite{liu2021one} unified MultiMF with partition generation to improve the clustering performance and efficiency for large-scale datasets. Graph-based methods strive to learn a unified graph from multiple views to delineate the pairwise similarities among data points. Huang et al.~\cite{huang2021measuring} proposed to formulate both the multi-view consistent graph and diverse graph in a unified framework. In~\cite{wang2021fast}, Wang et al. performed both anchor learning and graph construction to acquire an anchor graph to promote clustering efficiency. Another category is subspace-based MVC methods, which aim to infer latent representations from different views in a shared subspace. One representative example is the work of LMSC~\cite{zhang2018generalized}. It integrated multiple views into a comprehensive latent representation subspace that encodes complementary information across different views. Unlike LMSC, Sun et al.~\cite{sun2021scalable} combined anchor learning and graph construction into a unified subspace to get a more discriminative clustering structure. Recently, several deep MVC approaches~\cite{10509800,qin2021semi,9975265,wang2021consistent} have also emerged to bolster clustering performance by harnessing the robust feature learning capabilities of deep neural networks.

Although these MVC methods have achieved good results, they fail to fully explore the high-order correlations between each view. To solve this problem, tensor-based MVC methods have developed recently~\cite{zhang2020tensorized,8896047,9531538,10221718,10177207,9806306}, which usually stack the similarity graphs from all views into a three-order tensor to capture the cross-view correlations with tensor-based regularization. For instance, Zhang et al.~\cite{zhang2020tensorized} integrated self-expression based subspace representations from various views into a low-rank tensor, capturing the global structure and exploring correlations across multiple views. To obtain a more effective tensor for clustering, in addition to just considering the global structure and high-order correlations, Qin et al. made further strides by proposing two works: they investigated the local structures of similarity matrices from different views using the Markov chain\cite{10221718}, and explored pairwise correlations based on the reconstruction of a shared similarity matrix~\cite{10177207}. Additionally, To exploit the relationship between low-rank tensors and label indicator matrices, Fu et al.~\cite{9806306} unified low-rank tensor learning with spectral embedding into a framework. However, while these methods have shown success in capturing high-order correlations, their practical application is hindered by the stacking of large similarity graphs learned from multiple views into a tensor. This process incurs significant storage and computational complexity, making it unscalable for large datasets. Although some methods~\cite{chen2023tensor,QinPW24} tried to address this problem by introducing bipartite graphs, they may result in sub-optimal solutions caused by an unstable anchor selection process. Furthermore, they often overlook the adverse effects of non-semantic information in each view, leading to the entanglement of latent semantic-unrelated and semantic-related features. As a consequence, the clustering performance may be compromised.

To address the aforementioned challenges, we propose a novel approach called fast Disentangled Slim Tensor Learning (DSTL) for multi-view clustering. Specifically, Instead of concentrating on the multi-view graph structures, the DSTL approach directly leverages the matrix factorization technique to obtain low-dimensional features. Inspired by Robust Principal Component Analysis (RPCA)~\cite{candes2011robust}, we disentangle the features of each view to learn both semantic-unrelated and semantic-related representations. Subsequently, we construct two slim tensors that not only mitigate the adverse effects of latent semantic-unrelated information but also capture high-order consistency among multiple views. The semantic-unrelated slim tensor is assumed sparse with $\ell_1$-norm regularization, while the semantic-related slim tensor is assumed low-rank with tensor nuclear norm regularization. Additionally, we incorporate a consensus alignment indicator matrix to align semantic-related representations across views, guiding the disentanglement of latent features. In summary, we present the contributions of our model as follows:
\begin{itemize}
	\item We propose a novel fast MVC approach named DSTL, which combines latent multi-view features disentanglement for learning semantic representations and slim tensor learning for capturing cross-view correlations in a unified framework.
	\item DSTL directly explores the high-order correlations among multi-view semantic-related representations via slim tensor learning. Latent feature disentanglement is adopted to mitigate the adverse effects of view-specific semantic-unrelated information.
	\item The semantic-related representations are aligned across views to guide the disentanglement through a consensus alignment indicator. Experiments demonstrate the effectiveness and efficiency of DSTL compared to various state-of-the-art MVC methods.
\end{itemize}

\begin{figure*}[t]
	\center
	\includegraphics[width=17.8cm,height=8.5cm]{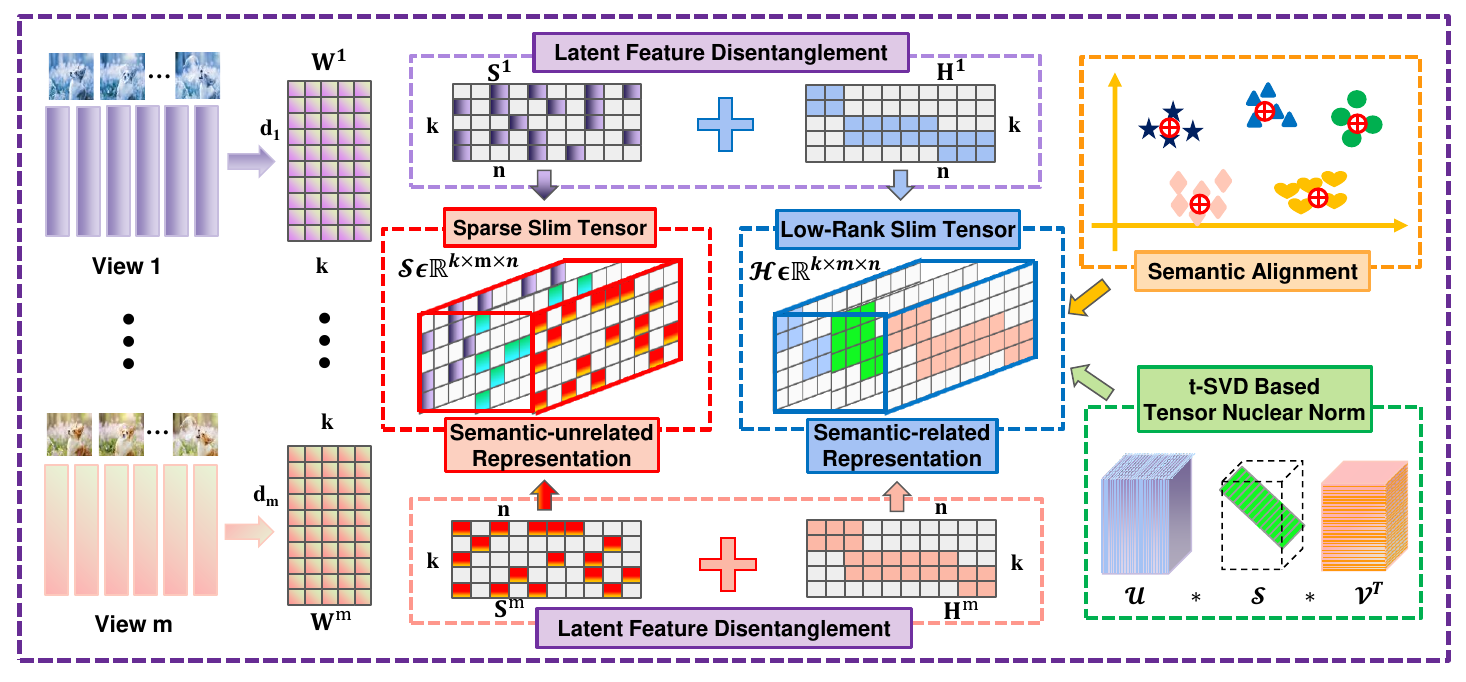}
	\caption{The overall framework of DSTL.}
	\label{framework}
\end{figure*}
\section{Notations and Related Work}
In this section, we explain the commonly used notations and provide necessary tensor preliminaries. Then we present two central contents related to our work: multi-view matrix factorization and tensor-based multi-view clustering.

\subsection{Notations and Preliminaries}
We adopt the following notation conventions throughout this paper: bold lowercase letters (e.g., $\mathbf{x}$) represent vectors, uppercase letters (e.g., $\mathbf{X}$) denote matrices, and calligraphic letters (e.g., $\mathcal{X}$) signify tensors.
For a matrix $\mathbf{X}\in \mathbb{R}^{n_1\times n_2}$, its Frobenius norm and nuclear norm are defined as $\|\mathbf{X}\|_F = \sqrt{\sum_{ij}x_{ij}^2}$ and $\|\mathbf{X}\|_* = \sum_{i}\delta_i(\mathbf{X})$, respectively, where $x_{ij}$ is the element of $\mathbf{X}$ at position ($i,j$), and $\delta_i(\mathbf{X})$ is the $i$-th singular value of $\mathbf{X}$. Tr($\cdot$) denotes the matrix trace, $\mathbf{1}$ represents an all-one vector, and $\mathbf{I}_k$ is a $k$-dimensional identity matrix. $sign(\cdot)$ denotes the sign of the input. For a tensor $\mathcal{X}\in \mathbb{R}^{n_1\times n_2 \times n_3}$, its $\ell_1$-norm is defined as $\|\mathcal{X}\|_1=\sum_{ijk}|x_{ijk}|$, where $x_{ijk}$ is the element of $\mathcal{X}$ at position ($i,j,k$). The $i$-th frontal, lateral, and horizontal slice of $\mathcal{X}$ are denoted as $\mathcal{X}^{(:,:,i)}$, $\mathcal{X}^{(:,i,:)}$, and $\mathcal{X}^{(i,:,:)}$, respectively. Additionally, for convenience, we use $\mathcal{X}^{(i)}$ to represent $\mathcal{X}^{(:,:,i)}$. $\mathcal{X}_{f}$ denotes the
fast Fourier transformation (FFT) of $\mathcal{X}$ along the third dimension, i.e., $\mathcal{X}_{f} = \text{fft}(\mathcal{X}, [], 3)$, and $\mathcal{X}$ can be recovered from $\mathcal{X}_{f}$ by the inverse FFT operation, i.e., $\mathcal{X} = \text{ifft}(\mathcal{X}_{f}, [], 3)$~\cite{lu2020tensor}.

\begin{definition}[\textbf{t-SVD}~\cite{kilmer2013third}]\label{def1}
	For a tensor $\mathcal{X}\in \mathbb{R}^{n_1\times n_2 \times n_3}$, its t-SVD is defined as
	$$
	\mathcal{X} = \mathcal{U}*\mathcal{S}*\mathcal{V}^T,
	$$
	where $\mathcal{U}\in \mathbb{R}^{n_1\times n_1 \times n_3}$ and $\mathcal{V}\in \mathbb{R}^{n_2\times n_2 \times n_3}$ are orthogonal
	tensors, $\mathcal{S}\in \mathbb{R}^{n_1\times n_2\times n_3}$ is an f-diagonal tensor, whose frontal slices are all diagonal matrices, and "$*$" denotes the t-product. 
\end{definition}

\begin{definition}[\textbf{t-SVD based tensor nuclear norm}~\cite{semerci2014tensor}]\label{def3}
	Given a tensor $\mathcal{X}\in \mathbb{R}^{n_1\times n_2 \times n_3}$, its t-SVD based tensor nuclear norm is defined as
	$$
	\|\mathcal{X}\|_{\circledast} = \sum_{k=1}^{n_3}\|\mathbf{X}^{(k)}_f\|_*=\sum_{i=1}^{\min(n_1,n_2)}\sum_{k=1}^{n_3}\delta_i(\mathbf{X}^{(k)}_f),
	$$
\end{definition}

\subsection{Multi-view Matrix Factorization}
Given the data from $m$ views $\{\mathbf{X}^v\}_{v=1}^m$, in which $\mathbf{X}^v \in \mathbb{R}^{d_v \times n}$ represents the $v$-th view with $d_v$ feature dimensions and $n$ samples. MultiMF tries to obtain a consensus latent representation $\mathbf{G}$ for all distinct views. The general objective function of MultiMF can be described as
\begin{equation}\label{MF}
\begin{aligned}
\min_{\mathbf{W}^{v},\mathbf{G}^{v}} &\sum_{v=1}^{m}\|\mathbf{X}^{v}-\mathbf{W}^{v}\mathbf{G}^{v}\|_F^2+\alpha\mathcal{T}\ (\mathbf{W}^v)+\lambda\mathcal{R}\ (\mathbf{G}, \mathbf{G}^v).\\
\end{aligned}
\end{equation}
Here, $\mathbf{W}^v\in\mathbb{R}^{d_v \times k}$ represents the view-specific base matrices. $\mathbf{G}^v \in \mathbb{R}^{k \times n}$ denotes the view-specific latent representation, and $\mathbf{G}$ is the consensus representation by fusing all $\{\mathbf{G}^v\}_{v=1}^m$. $k$ is the dimension of latent representation. $\alpha$ and $\lambda$ are the trade-off hyperparameters, $\mathcal{T}(\cdot)$ and $\mathcal{R}(\cdot)$ indicate certain regularization terms. For instance, Liu et al.~\cite{liu2013multi} used nonnegative constraints to compel clustering solutions of every view toward a shared consensus. Gao et al.~\cite{gao2019multi} utilized the Laplacian matrix of each view to capture the intrinsic data structure, alongside the consensus coefficient representation $\mathbf{G}$ in all data views. In \cite{Wan2023AWMVC}, Wan et al. introduced orthogonality constraints to improve the discriminative power of representation matrices. However, although Eq.~(\ref{MF}) can efficiently obtain the latent representation, it lacks the exploration of high-order cross-view correlations.

\subsection{Tensor-based Multi-view Clustering}


Tensor-based MVC methods seek to harness the high-order correlations among views by merging affinity matrices from all views into a three-order tensor, and then imposing tensor-level constraints to capture the inter-relationship among views. Given the data from $m$ views $\{\mathbf{X}^v\}_{v=1}^m$, tensor-based MVC methods usually take the following form:
\begin{equation}\label{Tensor}
\begin{aligned}
\min_{\mathbf{Z}^{v}} &\|\mathcal{Z}\|_{\circledast}+\gamma\mathcal{F}\ (\mathbf{X}^{v},\mathbf{Z}^{v})
&s.t.~\mathcal{Z} = \Phi(\mathbf{Z}^{1},...,\mathbf{Z}^{m}),\
\end{aligned}
\end{equation}
where $\gamma$ is the trade-off hyperparameter. $\mathcal{F}(\cdot)$ is to adaptively learn view-specific similarity graphs $\mathbf{Z}^{v} \in \mathbb{R}^{n \times n}$ from $\mathbf{X}^{v}$.  For instance, Wang et al.\cite{wang2023robustness} utilized a self-representation method to learn $\mathbf{Z}^v$ (e.g., $\mathbf{X}^v=\mathbf{X}^v\mathbf{Z}^v$), while Chen et al.\cite{chen2023low} learned the similarity graph based on the distances between data points (e.g., $\|\mathbf{x}^v_i-\mathbf{x}^v_j\|_2^2{z}^v_{ij}$). $\Phi(\cdot)$ denotes merging and rotating $\mathbf{Z}^v$ into a three-order tensor $\mathcal{Z}^v \in \mathbb{R}^{n\times m\times n}$. $\|\mathcal{Z}\|_{\circledast}$ is the tensor nuclear norm which constrains low-rank attributes on $\mathbf{Z}^{v}$ and defined by Definition~\ref{def3}.

However, when stacking the similarity graph $\mathbf{Z}^v$ as a tensor, it will take $O(n^2m)$ storage complexity, and tensor-related operations such as FFT and t-SVD usually have an estimated time complexity of $O(n^2log(n))$, limiting the application on large-scale datasets. Moreover, they directly explore the correlations among multi-view similarity graphs and neglect the semantic-unrelated information in each view, which can lead to a performance decrease.

\section{The Proposed Method}
To address the above challenges, we propose a novel fast MVC approach called DSTL in this section. Then, we present the optimization algorithm and analyze the algorithm's computational complexity. The overall framework of our DSTL is shown in Fig. \ref{framework}.
\subsection{Model Formulation}
In modern multi-view applications, where data tends to be both massive (i.e., $n_v$ is very large) and high-dimensional (i.e., $d_v$ is very large), posing a great challenge to efficient data clustering analysis~\cite{liu2016blessing}. Similarity graph based methods usually explore the similarity relations between pairwise data points for spectral clustering, making them unscalable to large data. Matrix factorization provides an effective solution to it, which directly obtains the latent low-dimensional representation for downstream tasks. As illustrated in Eq.~(\ref{MF}), most previous MultiMF based approaches learn a latent representation for each view and fuse them into a single one. However, such a strategy ignores the potentially irrelevant information in each view (e.g., irreverent backgrounds in the visual view, non-meaningful words in the textual view and occlusion on human faces~\cite{zhang2020enhanced}), which may degrade the quality of the final representation. We call those irrelevant information semantic-unrelated information. And it is natural to expect the elimination of its negative influence. In this paper, we employ feature disentanglement to remove semantic-unrelated information and seek the intrinsic semantic-related information. The objective function can be formulated as

\begin{equation}\label{MFdisentangle}
\begin{aligned}
\min_{\mathbf{W}^{v},\mathbf{S}^{v},\mathbf{H}^{v}} &\sum_{v=1}^{m}\|\mathbf{X}^{v}-\mathbf{W}^{v}(\mathbf{S}^{v}+\mathbf{H}^{v})\|_F^2\\
s.t.&~\mathbf{W}{^{v^T}}\mathbf{W}^{v}=\mathbf{I}_k,
\end{aligned}
\end{equation}
where $\mathbf{S}^v\in \mathbb{R}^{k\times n}$ and $\mathbf{H}^v\in \mathbb{R}^{k\times n}$ denote the latent semantic-unrelated representation and semantic-related representation of the $v$-th view, respectively. The base matrix $\mathbf{W}^v$ is imposed orthogonality constraints to avoid arbitrary scaling. 

It can be observed that there are infinite disentanglement results w.r.t. $\mathbf{S}^v$ and $\mathbf{H}^v$ in Eq. (\ref{MFdisentangle}). Thus, it is necessary to impose constraints or regularization on the two parts to achieve a meaningful disentanglement. RPCA~\cite{candes2011robust} is a classical and effective data recovery approach. It decomposes the original data into the sum of a sparse component and a low-rank component, which characterizes the noise and intrinsic data, respectively. It can theoretically guarantee that the original true data can be exactly recovered and principal features extracted under certain mild conditions. In Eq. (\ref{MFdisentangle}), the original latent representation $\mathbf{G}^v$ is decomposed into a semantic-unrelated part and an intrinsic semantic-related part. Inspired by RPCA, we reformulate Eq.~(\ref{MFdisentangle}) as follows by sparse ($\ell_1$-norm) and low-rank (nuclear norm) regularization to enhance the latent representation disentanglement.
\begin{equation}\label{RPCA}
	\begin{aligned}
\min_{\mathbf{W}^{v},\mathbf{S}^{v},\mathbf{H}^{v}} &\sum_{v=1}^{m}\|\mathbf{X}^{v}-\mathbf{W}^{v}(\mathbf{S}^{v}+\mathbf{H}^{v})\|_F^2  \\
&+\lambda_1\|\mathbf{S}^v\|_{1}+\lambda_2\|\mathbf{H}^v\|_{\ast}\\
s.t.&~\mathbf{W}{^{v^T}}\mathbf{W}^{v}=\mathbf{I}_k.
	\end{aligned}
\end{equation}

However, Eq.~(\ref{RPCA}) still have some limitations. Firstly, it treats each view individually, and ignores the high-order cross-view correlations. Secondly, although it seeks the intrinsic low-rank representations for each view, the semantic-consistency across views among these representations is not guaranteed. To address the first problem, inspired by the success of tensor-based regularization, we stack and rotate the disentangled low-dimensional features $\mathbf{S}^{v}$ and $\mathbf{H}^{v}$ into two slim tensors $\mathcal{S} \in \mathbb{R}^{k \times m \times n}$ and $\mathcal{H} \in \mathbb{R}^{k \times m \times n}$. Then, we employ $\ell_1$-norm and t-SVD based nuclear norm regularization terms on $\mathcal{S}$ and $\mathcal{H}$, which encourages sparsity and low-rank properties of the semantic-unrelated and semantic-related representations, respectively. Then, Eq. (\ref{RPCA}) becomes
\begin{equation}\label{Slim tensor}
	\begin{aligned}
		\min_{\mathbf{W}^{v},\mathcal{S},\mathcal{H}} &\sum_{v=1}^{m}\|\mathbf{X}^{v}-\mathbf{W}^{v}(\mathbf{S}^{v}+\mathbf{H}^{v})\|_F^2+ \lambda_1\|\mathcal{S}\|_{1} +\lambda_2\|\mathcal{H}\|_{\circledast}\\
		s.t.&~\mathbf{W}{^{v^T}}\mathbf{W}^{v}=\mathbf{I}_k,\\
		&~\mathcal{S} = \Phi(\mathbf{S}^{1},...,\mathbf{S}^{m}),~\mathcal{H} = \Phi(\mathbf{H}^{1},...,\mathbf{H}^{m}).\
	\end{aligned}
\end{equation}
Through this formulation, we encourage the semantic-unrelated representation to be a sparse slim tensor and effectively filter out noise and irrelevant features. Simultaneously, the semantic-related representation are encouraged to be a low-rank slim tensor, which effectively capturing underlying structures and meaningful patterns as well as high-order correlations across multiple views.

For the second problem, we address it by aligning the data semantics of different views. It is known that the features of a same object belongs to a same cluster in all views. Considering that the latent semantic-related features $\mathbf{H}^v$ should contain the essential information reflecting the belonging cluster, we learn a consensus alignment indicator $\mathbf{Y} \in \mathbb{R}^{k \times n}$ to align semantic-related representations across all views, and it also reversely guide the process of feature disentanglement, i.e., 
\begin{equation}\label{Indicator}
\begin{aligned}
\min_{\mathbf{C}^{v},\mathbf{Y}} &\sum_{v=1}^{m}\|\mathbf{H}^{v}-\mathbf{C}^{v}\mathbf{Y}\|_F^2\\
s.t.&~\mathbf{C}{^{v^T}}\mathbf{C}^{v}=\mathbf{I}_k, \mathbf{Y} \geq 0, \mathbf{Y}^T\mathbf{1}=\mathbf{1},\
\end{aligned}
\end{equation}
where $\mathbf{C}^{v} \in \mathbb{R}^{k \times k}$ is the semantic center matrix of the $v$-th view. We adopt constraints $\mathbf{Y} \geq 0$ and $\mathbf{Y}^T\mathbf{1}=\mathbf{1}$ to signify the probability of each sample in semantic-related representations being assigned to a specific semantic center.

By integrating Eqs.~(\ref{Slim tensor}), and~(\ref{Indicator}), the objective function of DSTL can be described as 
\begin{equation}\label{Objective}
\begin{aligned}
\min_{\Omega} &\sum_{v=1}^{m}\|\mathbf{X}^{v}-\mathbf{W}^{v}(\mathbf{S}^{v}+\mathbf{H}^{v})\|_F^2+ \lambda_1\|\mathcal{S}\|_{1} \\
&+\lambda_2\|\mathcal{H}\|_{\circledast}+\lambda_3\sum_{v=1}^{m}\|\mathbf{H}^{v}-\mathbf{C}^{v}\mathbf{Y}\|_F^2\\
s.t.&~\mathbf{W}{^{v^T}}\mathbf{W}^{v}=\mathbf{I}_k,\mathbf{C}{^{v^T}}\mathbf{C}^{v}=\mathbf{I}_k, \mathbf{Y} \geq 0, \mathbf{Y}^T\mathbf{1}=\mathbf{1},\\
&~\mathcal{S} = \Phi(\mathbf{S}^{1},...,\mathbf{S}^{m}),~\mathcal{H} = \Phi(\mathbf{H}^{1},...,\mathbf{H}^{m}), \
\end{aligned}
\end{equation}
where $\Omega=\{\mathbf{W}^v, \mathcal{S},\mathcal{H}, \mathbf{C}^v, \mathbf{Y}\}$ is the target variables set. $\lambda_1$, $\lambda_2$ and $\lambda_3$ are the trade-off hyper-parameters. As can be observed, different from previous tensor-based MVC approaches that focus on the similarity structure exploration, our model directly learns the high-order correlated multi-view latent representations via matrix factorization and slim tensor regularization. Moreover, our method also mitigates the negative influences of view-specific semantic-unrelated information by RPCA induced feature disentanglement. A consensus semantic alignment indicator is further employed to capture the semantic-consistency across views and also enhance the quality of latent feature disentanglement. After solving Eq. (\ref{Objective}), we can apply $k$-means on $\mathbf{Y}$ to obtain the clusters.

\subsection{Optimization}
To optimize the objective function involving multiple variables, we adopt an alternate optimization strategy. Specifically, we update one variable in each step while keeping the others fixed.

\textbf{Updating} $\mathbf{W}^v$: When other variables are fixed, the optimization problem w.r.t. $\mathbf{W}^v$ can be formulated as follows:
\begin{equation}\label{W}
\begin{aligned}
\mathbf{W}^v_{t+1} = \argmin_{\mathbf{W}^{v}} &\sum_{v=1}^{m}\|\mathbf{X}^{v}-\mathbf{W}^{v}(\mathbf{S}^{v}_t+\mathbf{H}^{v}_t)\|_F^2\\
s.t.&~{\mathbf{W}^{v^T}}\mathbf{W}^{v}=\mathbf{I}_k.\
\end{aligned}
\end{equation}
By transforming the Frobenius norm to the trace and eliminating terms irrelevant to $\mathbf{W}^v$, the above formula can be equivalently reformulated as:
\begin{equation}\label{TrW}
	\begin{aligned}
		\max_{\mathbf{W}^v}
		~&\mathrm{Tr}({\mathbf{W}^{v^T}}\mathbf{M}^v_t)
		&s.t.~\mathbf{W}{^{v^T}}\mathbf{W}^{v}=\mathbf{I}_k,\
	\end{aligned}
\end{equation}
$\mathbf{M}^v_t=\mathbf{X}^v(\mathbf{S}^{v}_t+\mathbf{H}^{v}_t)^T$. This subproblem can be efficiently solved by Theorem~\ref{SVD}.
\begin{theorem}\label{SVD}
Let $\mathbf{B}=\mathbf{U}\mathbf{\Sigma}\mathbf{V}^T$ be the singular value decomposition (SVD) of a matrix $\mathbf{B}$. The optimal solution to the problem
\begin{equation}\label{T1}
	\begin{aligned}
		\max_{\mathbf{A}^v}
		~&\mathrm{Tr}({\mathbf{A}}\mathbf{B})
		&s.t.~\mathbf{A}\mathbf{A}^T=\mathbf{I},\
	\end{aligned}
\end{equation}
is given by $\mathbf{B}^\dagger=\mathbf{V}\mathbf{U}^T$.
\end{theorem}
\noindent\textit{Proof}. Substituting $\mathbf{B}=\mathbf{U}\mathbf{\Sigma}\mathbf{V}^T$ into Eq.~(\ref{T1}), we get
\begin{equation}\label{T2}
	\begin{aligned}
        \mathrm{Tr}({\mathbf{A}}\mathbf{B})&=\mathrm{Tr}({\mathbf{A}}\mathbf{U}\mathbf{\Sigma}\mathbf{V}^T)=\mathrm{Tr}(\mathbf{\Sigma}\mathbf{V}^T\mathbf{A}\mathbf{U})\\
        &=\mathrm{Tr}(\mathbf{\Sigma}\mathbf{O})=\sum_{i}s_{ii}o_{ii},
	\end{aligned}
\end{equation}
where $\mathbf{O}=\mathbf{V}^T\mathbf{A}\mathbf{U}$. It can be easily verified that $\mathbf{O}\mathbf{O}^T=\mathbf{I}$. Thus, we have $-1 \leq o_{ii} \leq 1$, and then
\begin{equation}\label{T3}
	\begin{aligned}
		\mathrm{Tr}({\mathbf{A}}\mathbf{B})=\sum_{i}s_{ii}o_{ii} \leq \sum_{i}s_{ii}.
	\end{aligned}
\end{equation}
Without of generality, let $\mathbf{O}=\mathbf{V}^T\mathbf{A}\mathbf{U}=\mathbf{I}$, then $\mathbf{B}^\dagger=\mathbf{V}\mathbf{U}^T$, and $\mathrm{Tr}({\mathbf{A}}\mathbf{B})$ attains its maximum. This completes the proof.\qed

\textbf{Updating} $\mathbf{C}^v$: When other variables are fixed, the optimization problem w.r.t. $\mathbf{C}^v$ is
\begin{equation}\label{C}
\begin{aligned}
\mathbf{C}^v_{t+1} = \argmin_{\mathbf{C}^{v}} &\sum_{v=1}^{m}\|\mathbf{H}^{v}_t-\mathbf{C}^{v}\mathbf{Y}_t\|_F^2
&s.t.~\mathbf{C}{^{v^T}}\mathbf{C}^{v}=\mathbf{I}_k.\
\end{aligned}
\end{equation}
Similar to the $\mathbf{W}^v$ problem, Eq.~(\ref{C}) is converted to
\begin{equation}\label{TrC}
	\begin{aligned}
		\max_{\mathbf{C}^v}
		~&\mathrm{Tr}({\mathbf{C}^{v^T}}\mathbf{N}^v_t)
		&s.t.~\mathbf{C}{^{v^T}}\mathbf{C}^{v}=\mathbf{I}_k,\
	\end{aligned}
\end{equation}
where $\mathbf{N}^v_t= \mathbf{H}^{v}_t\mathbf{Y}^{T}_t$. As in solving Eq.~(\ref{TrW}), the optimal solution for $\mathbf{C}^v$ can be obtained by Theorem~\ref{SVD}.

\textbf{Updating} $\mathcal{S}$: When other variables are fixed, the optimization problem w.r.t. $\mathcal{S}$ is

\begin{equation}\label{S}
\begin{aligned}
\mathcal{S}_{t+1} = \argmin_{\mathcal{S}}\lambda_1\|\mathcal{S}\|_1+\| \mathcal{S}-\mathcal{P}_{t}\|^2_F,
\end{aligned}
\end{equation}
where $\mathcal{P}_{t} = \Phi(\mathbf{P}^{1}_{t},...,\mathbf{P}^{m}_{t})$ is constructed by
\begin{equation}\label{P}
	 \mathbf{P}^{v}_{t} = \mathbf{W}{^{v^T}_{t+1}}\mathbf{X}^{v}-\mathbf{H}^{v}_t.
\end{equation} 
Eq.~(\ref{S}) can be solved by

\begin{equation}\label{sf}
\begin{aligned}
\mathbf{S}^{v}_{t+1} = \mathbb{D}_{\frac{\lambda_1}{2}}(\mathbf{P}^{v}_{t}),
\end{aligned}
\end{equation}
where $\mathbb{D}_{\gamma}$ is the soft-threshold operator~\cite{yang2011alternating} defined as $[\mathbb{D}_{\gamma}(\mathbf{A})]_{ij} = sign\ (a_{ij}) \cdot \{\max(|a_{ij}|-\gamma,0)\}$.

\textbf{Updating} $\mathcal{H}$: When other variables are fixed, we can update $\mathcal{H}$ by working out the following subproblem

\begin{equation}\label{H}
\begin{aligned}
\mathcal{H}_{t+1}=\argmin_{\mathcal{H}}\lambda_2\|\mathcal{H}\|_{\circledast}+(\lambda_3+1)\|\mathcal{H}-\mathcal{Q}_{t}\|^2_F,
\end{aligned}
\end{equation}
where $\mathcal{Q}_{t} = \Phi(\mathbf{Q}^{1}_{t},...,\mathbf{Q}^{m}_{t})$ is constructed using the following equation:

\begin{equation}\label{Q}
	\begin{aligned}
    \mathbf{Q}^{v}_{t} = \frac{1}{\lambda_3+1}(\mathbf{W}{^{v^T}_{t+1}}\mathbf{X}^{v}-\mathbf{S}^{v}_{t+1})+\frac{\lambda_3}{\lambda_3+1}\mathbf{C}{^{v^T}_{t+1}}\mathbf{Y}_{t}. 
\end{aligned}
\end{equation}

Eq.~(\ref{H}) is a typical low-rank tensor norm minimization problem that has a closed-form solution and can be easily solved by the following Theorem~\ref{tensor}~\cite{xie2018unifying}.
\begin{theorem}\label{tensor}
	Given two tensor $\mathcal{K}\in \mathbb{R}^{n_1\times n_2 \times n_3}$ and $\mathcal{L}\in \mathbb{R}^{n_1\times n_2 \times n_3}$ with a constant $\rho$, the globally optimal solution of the problem
	$$
	\min_{\mathcal{K}}\rho\|\mathcal{K}\|_{\circledast}+{\frac{1}{2}}\|\mathcal{K}-\mathcal{L}\|^2_F
	$$
	can be obtained by the tensor tubal-shrinkage operator
	$$
	\mathcal{K}=\mathcal{C}_{n_3\rho}(\mathcal{L})=\mathcal{U}*\mathcal{C}_{n_3\rho}(\mathcal{Z})*\mathcal{V}^T,
	$$
	where $\mathcal{L}=\mathcal{U}*\mathcal{Z}*\mathcal{V}^T $ and $\mathcal{C}_{n_3\rho}(\mathcal{L})=\mathcal{Z}*\mathcal{D}$. $\mathcal{D}\in \mathbb{R}^{n_1\times n_2 \times n_3}$ is a $f$ -diagonal tensor and its diagonal element in the Fourier domain is $\mathcal{D}_f(i,i,j)=(1-\frac{n_3\rho}{\mathcal{Z}_{f}^{(j)}(i,i)})_+$.
\end{theorem}
\begin{algorithm}[tb]
	\caption{DSTL algorithm}
	\label{alg:algorithm}
	\textbf{Input}: \hfill
	\parbox[t]{0.86\linewidth}{Multi-view data matrices $\{\mathbf{X}^{v}\}_{v=1}^m$, parameters $\lambda_1, \lambda_2, \lambda_3$, and $k$.}\\
	
	\textbf{Output}: \hfill
	\parbox[t]{0.86\linewidth}{Perform \textit{k}-means on \textbf{Y}.}\\
	\vspace{-\baselineskip}
	\begin{algorithmic}[1] 
		\STATE Initialize $\mathbf{W}^{v}=\mathbf{0}$, $\mathbf{C}^{v}=\mathbf{0}$, $\mathbf{Y}=\mathbf{0}$, $\mathcal{S} = \mathcal{H} =0$, $\epsilon=1e-4$. 
		\WHILE{not converged}
		\STATE Update $\mathbf{W}^v$ by solving~(\ref{W});
		\STATE Update $\mathbf{C}^v$ by solving~(\ref{C});
		\STATE Update $\mathcal{S}$ by Eq.~(\ref{sf});
		\STATE Update $\mathcal{H}$ by solving.~(\ref{H});
		\STATE Update $\mathbf{Y}$ by solving~(\ref{sY});
		\STATE Check the convergence conditions:\\ $\|\mathbf{Y}_{t}-\mathbf{Y}_{t-1}\|_F^2/\|\mathbf{Y}_{t-1}\|_F^2\le \epsilon$ ;
		\STATE $t \longleftarrow t+1$;
		\ENDWHILE
		\STATE \textbf{return} The alignment indicator matrix $\mathbf{Y}$.
	\end{algorithmic}
\end{algorithm}
\textbf{Updating} $\mathbf{Y}$: When other variables are fixed, the optimization problem w.r.t. $\mathbf{Y}$ is
\begin{equation}\label{Y}
\begin{aligned}
\mathbf{Y}_{t+1} = \argmin_{\mathbf{Y}} &\sum_{v=1}^{m}\|\mathbf{C}{^{v^T}_{t+1}}\mathbf{H}^{v}_{t+1}-\mathbf{Y}\|_F^2\\
s.t.&~\mathbf{Y} \geq 0, \mathbf{Y}^T\mathbf{1}=\mathbf{1}.
\end{aligned}
\end{equation}
Because the consensus alignment vectors are independent with each other, $\mathbf{y}_i$ can be solved in a column-wise manners, i.e.,
\begin{equation}\label{sY}
	\begin{aligned}
		\|\mathbf{f}_i-\mathbf{y}_i\|_2^2
		~~s.t.~\mathbf{y}_i \geq 0, \mathbf{y}_i^T\mathbf{1}=1,
	\end{aligned}
\end{equation}
where $\mathbf{f}_i$ is the vetor of $\mathbf{F}=\sum_{v=1}^{m}\mathbf{C}{^{v^T}_{t+1}}\mathbf{H}^{v}_{t+1}$. It can be efficiently solved by Theorem~\ref{solveY}~\cite{huang2015new}.
\begin{theorem}\label{solveY}
	Given a vector $\mathbf{g}\in \mathbb{R}^n$ the following problem has a closed-form solution $\mathbf{y}^\dagger$, i.e.,
	$$
    \mathbf{y}^\dagger=\argmin\frac{1}{2}\|\mathbf{g}-\mathbf{y}\|_2^2~s.t.~\mathbf{y} \geq 0, \mathbf{y}^T\mathbf{1}=1.
	$$
\end{theorem}
\noindent\textit{Proof}. We start by defining the Lagrangian function:
\begin{equation}\label{P1}
	\begin{aligned}
      \mathcal{L}(\mathbf{y},\bm{\xi},\rho)=\frac{1}{2}\|\mathbf{g}-\mathbf{y}\|_2^2-\bm{\xi}^T\mathbf{y}-\rho(\mathbf{y}^T\mathbf{1}-1),
	\end{aligned}
\end{equation}
where $\bm{\xi}$ is a Lagrangian coefficient vector and $\rho$ is a scalar. The optimal solution of $\mathbf{y}$ satisfy the Karush-Kuhn-Tucker (KKT) conditions that:
\begin{equation}\label{P2}
	\forall i:
	\left\{
	\begin{aligned}
		& y_i - g_i - \xi^\dagger_i - \rho^\dagger = 0, \\
		& y_i \geq 0, \mathbf{y}^T\mathbf{1}=1, \\
		& \xi_i^\dagger \geq 0, \xi_i^\dagger y_i=0,
	\end{aligned}
	\right.
\end{equation}
where $\xi^\dagger_i$ and $\rho^\dagger$ are the optimal Lagrangian multipliers. By solving Eq.~(\ref{P1}), the optimal solution is $y_i=(g_i+\rho^\dagger)_+$, here $(a)_+=max(a,0)$. Due to $\mathbf{y}^T\mathbf{1}=1$ and $\xi_i^\dagger \geq 0$, we have
\begin{equation}\label{P3}
	\left\{
	\begin{aligned}
		& \rho^\dagger = v_i-g_i-\tilde{\xi}^\dagger, \\
		& y_i=v_i-\tilde{\xi}^\dagger+\xi^\dagger, \\
	\end{aligned}
	\right.
	\Longrightarrow y_i=(v_i-\tilde{\xi}^\dagger)_+,
\end{equation}
where $v_i=g_i-(\mathbf{g}^T\mathbf{1})/n+1/n$ and $\tilde{\xi}^\dagger=({\xi_i^\dagger}^T\mathbf{1})/n$. Based on Eq.~(\ref{P3}) and $y_i \geq 0$, we can obtain
\begin{equation}\label{P4}
	\begin{aligned}
		& \xi_i^\dagger = (\tilde{\xi}^\dagger-v_i)_+ \\
	\end{aligned}
	\Longrightarrow \tilde{\xi}^\dagger=\frac{1}{n}\sum_{i=1}^{n}(\tilde{\xi}^\dagger-v_i)_+.
\end{equation}
By introducing $f(\tilde{\xi}^\dagger)=(1/n)\sum_{i=1}^{n}(\tilde{\xi}^\dagger-v_i)_+-\tilde{\xi}^\dagger$, $\tilde{\xi}^\dagger$ would be achieved when $f(\tilde{\xi}^\dagger)=0$, which can be solved by the Newton's method
\begin{equation}\label{P5}
	\begin{aligned}
		\tilde{\xi}_{t+1}^\dagger=\tilde{\xi}_{t}^\dagger-\frac{f(\tilde{\xi}_{t}^\dagger)}{f^{'}(\tilde{\xi}_{t}^\dagger)},
	\end{aligned}
\end{equation}
where $t$ is the iteration number. Therefore, the close-form solution is given by Eq.~(\ref{P3}). This completes the proof.\qed

By updating these variables iteratively, we can obtain the consensus alignment indicator embedding $\mathbf{Y}$. Then we perform $k$-means on $\mathbf{Y}$ to obtain the final results. Algorithm \ref{alg:algorithm} summarizes the detailed optimization process.

\subsection{Computational Complexity Analysis}
\subsubsection{Space Complexity}
The major memory costs of our algorithm are variables $\mathbf{X}^v \in \mathbb{R}^{d_v \times n}$, $\mathbf{W}^v \in \mathbb{R}^{d_v \times k}$, $\mathbf{S}^v \in \mathbb{R}^{k \times n}$,  $\mathbf{H}^v \in \mathbb{R}^{k \times n}$, $\mathbf{C}^v \in \mathbb{R}^{k \times k}$, and $\mathbf{Y} \in \mathbb{R}^{k \times n}$. Therefore, the space complexity of Algorithm~\ref{alg:algorithm} is linear to the number of samples, i.e., $O(n)$.
\subsubsection{Time Complexity}
The most time-consuming steps for updating $\mathbf{W}^{v}$ and $\mathbf{C}^{v}$ are the SVD operation and matrix multiplication, which require $\mathcal{O}(k^2d_v+kd_vn)$ and $O(k^3+k^2n)$, respectively. When updating $\mathcal{S}$ and $\mathbf{Y}$, the primary time cost is matrix multiplication, which takes $O(kd_vn)$ and $O(k^2n)$, respectively. In terms of updating $\mathcal{H}$, we mainly consider the computational complexity of matrix multiplication, FFT, inverse FFT, and SVD operations. It takes $O(kd_vn+k^2n)$ in matrix multiplication. For a $k\times m \times n$ tensor, it takes $O(kmnlog(n))$ to conduct FFT and inverse FFT operations, and needs $O(knm^2)$ for SVD operation. Since $k$ and $m$ are much smaller than $n$ and $d_v$, the overall computational complexity of Algorithm \ref{alg:algorithm} is $O(\tau(kmnlog(n)+kdn))$, where $d=\sum_{v=1}^{m}d_v$, $\tau$ is the number of iterations. In addition, the post-processing of the \textit{k}-means step has linear complexity to $n$. However, most existing tensor-based MVC approaches often have $O(n^2log(n))$ at the FFT stage and $O(n^3)$ at the spectral clustering stage. Therefore, the DSTL reduces the complexity of tensor-based MVC methods and is scalable to large-scale datasets.

\subsection{Convergence Analysis}
When considering all the variables jointly, the overall objective function is non-convex, but the convergence of each sub-problem in Algorithm~\ref{alg:algorithm} can be guaranteed, and each closed-form solution reduces the objective function value. Let $\Omega(\mathbf{W}^{v}_t,\mathbf{C}^{v}_t,\mathbf{S}^{v}_t,\mathbf{H}^{v}_t,\mathbf{Y}_t)$ be the objective function value after the $t$-th iteration, and we can conclude $\Omega(\mathbf{W}^{v}_t,\mathbf{C}^{v}_t,\mathbf{S}^{v}_t,\mathbf{H}^{v}_t,\mathbf{Y}_t) \geq \Omega(\mathbf{W}^{v}_{t+1},\mathbf{C}^{v}_t, \mathbf{S}^{v}_t,\mathbf{H}^{v}_t,\mathbf{Y}_t) \geq \Omega(\mathbf{W}^{v}_{t+1},\mathbf{C}^{v}_{t+1},\mathbf{S}^{v}_t,\mathbf{H}^{v}_t, \mathbf{Y}_t)$ $\geq \Omega(\mathbf{W}^{v}_{t+1},\mathbf{C}^{v}_{t+1},\mathbf{S}^{v}_{t+1},\mathbf{H}^{v}_t,\mathbf{Y}_t)\geq  \Omega(\mathbf{W}^{v}_{t+1},\mathbf{C}^{v}_{t+1},\mathbf{S}^{v}_{t+1},$ $\mathbf{H}^{v}_{t+1},\mathbf{Y}_t) \geq \Omega(\mathbf{W}^{v}_{t+1},\mathbf{C}^{v}_{t+1},$ $\mathbf{S}^{v}_{t+1},\mathbf{H}^{v}_{t+1},\mathbf{Y}_{t+1})$. Since the overall objective function value is lower-bounded, it can converge to a local minimum eventually.

\section{Experiment}
\subsection{Experimental Setup}

\textbf{Datasets:} Nine popular datasets from diverse applications were used to validate our DSTL, consisting of text datasets \textbf{NGs}\footnote{https://lig-membres.imag.fr/grimal/data.html} and \textbf{BBCSport}, digit dataset \textbf{HW}\footnote{https://archive.ics.uci.edu/ml/datasets/Multiple+Features}, scene datasets \textbf{Scene15}~\cite{fei2005bayesian} and \textbf{MITIndoor}~\cite{quattoni2009recognizing}, video dataset \textbf{CCV}~\cite{wang2021fast}, animal dataset \textbf{Animal}\footnote{\url{https://github.com/wangsiwei2010/large_scale_multi-view_clustering_datasets}}, as well as object datasets \textbf{Caltech101-all}\footnote{https://data.caltech.edu/records/mzrjq-6wc02} and \textbf{NUSWIDEOBJ}~\cite{chua2009nus}. More details are shown in Table~\ref{tab:datasets}. 

\begin{table}[!t]\large
		\renewcommand{\arraystretch}{1.02}
	\centering
	\caption{Details of the used datasets.}
	\resizebox{\linewidth}{!}{
		\begin{tabular}{ccccc}
			\toprule
			\textbf{Dataset} & \textbf{Sample} & \textbf{Cluster} & \textbf{View} & \textbf{View dimension} \\
			\midrule
			\textbf{NGs} & 500   & 5     & 3     & 2000, 2000, 2000 \\
			\textbf{BBCSport} & 544   & 5     & 2     & 3183, 3203 \\
			\textbf{HW} & 2000  & 10    & 6     & 240, 76, 216, 47, 64, 6 \\
			\textbf{Scene15} & 4485  & 15    & 3     & 1800, 1180, 1240 \\
			\textbf{MITIndoor} & 5360  & 67    & 4     & 4096, 3600, 1770, 1240 \\
			\textbf{CCV} & 6773  & 20    & 3     & 20, 20, 20 \\
			\textbf{Caltech101-all} & 9144  & 102   & 5     & 48, 40, 254, 512, 928 \\
			\textbf{Animal} & 11673 & 20    & 4     & 2689, 2000, 2001, 2000 \\
			\textbf{NUSWIDEOBJ} & 30000 & 31    & 5     & 64, 225, 144, 73, 128 \\
			\bottomrule
		\end{tabular}%
	}
	
	\label{tab:datasets}%
\end{table}%

\textbf{Evaluation Metrics:} Five commonly used metrics are utilized to assess the performance of clustering, including accuracy (ACC), normalized mutual information (NMI), purity (PUR), adjusted rand index (ARI), and F-score. Higher values for all metrics indicate superior performance.

\textbf{Baselines:} We compare the clustering performance of DSTL against eight state-of-the-art MVC approaches:
\begin{enumerate}[]
\item \textbf{AWMVC}~\cite{Wan2023AWMVC} utilizes MultiMF, then maps data matrices into various latent spaces and merges coefficient matrices from different spaces into a consensus one.

\item \textbf{FDAGF}~\cite{zhang2023let} optimizes contribution weights of pre-defined anchor numbers and uses a novel hybrid fusion approach for multi-size anchor graphs.

\item \textbf{FSMSC}~\cite{chen2023fast} incorporates both view-shared anchor learning and global-guided-local self-guidance learning into a unified model for fast MVC. 

\item \textbf{UDBGL}~\cite{fang2023efficient} adopts anchor learning to learn view-specific and view-consensus bipartite graphs, as well as integrates discrete clusters into a unified framework.

\item \textbf{t-SVD-MSC}~\cite{xie2018unifying} employs a tensor nuclear norm in a cohesive tensor space to capture and propagate complementary information among all views.

\item \textbf{ETLMSC}~\cite{wu2019essential} constructs a tensor nuclear norm combined with a Markov chain to capture the principle information from multiple views.

\item \textbf{TBGL}~\cite{xia2022tensorized} utilizes spatial structure and complementary information within bipartite graphs of views by minimizing the tensor Schatten \textit{p}-norm.

\item \textbf{LTBPL}~\cite{chen2023low} stacks all probability affinity matrices into a low-rank tensor and links view-specific representation with consensus indicator graph learning.
\end{enumerate}
Among these baselines, AWMVC is the MultiMF-based, FDAGF, FSMSC and UDBGL are the anchor graph-based, and the last four baselines are the tensor-based methods.

\textbf{Parameter Settings:} The hyperparameters in the baselines are tuned as suggested by the corresponding papers. For our DSTL, we simply fix $\lambda_3=1e-4$ and $k=c$ ($c$ is the number of clusters) for all datasets. $\lbrace\lambda_1,\lambda_2\rbrace$ are tuned in the range of $\lbrace1e-4, 5e-4, 1e-3,..., 5e0\rbrace$. We run all methods 10 times to obtain clustering results and standard deviations for a fair comparison. All experiments are performed using MATLAB R2021a with i5-12400 CPU and 16GB RAM on a PC.

\begin{table*}[!t]\large
	\renewcommand{\arraystretch}{1.1}
	\centering
	\caption{Clustering results (mean(std)) of different methods on nine datasets. OM indicates \textit{out of memory}.}
		\resizebox{\linewidth}{!}{
			\begin{tabular}{c|c|ccccccccc}
				\hline
				\textbf{Dataset} & \textbf{Metric(\%)}&\textbf{AWMVC} & \textbf{FDAGF} & \textbf{FSMSC} & \textbf{UDBGL} & \textbf{t-SVD-MSC} & \textbf{ETLMSC} & \textbf{TBGL} & \textbf{LTBPL} & \textbf{Ours} \\
				\hline
				\multirow{5}[2]{*}{\textbf{NGs}} & \textbf{ACC} & 88.38(12.65) & 70.56(0.13) & 96.16(6.45) & 38.40(0) & \textbf{100(0)} & 67.18(0.38) & 59.80(0) & \underline{99.80(0)} & \textbf{100(0)} \\
				& \textbf{NMI} & 85.25(6.62) & 55.79(0.15) & 93.33(2.92) & 12.05(0) & \textbf{100(0)} & 58.11(0.92) & 53.06(0) & \underline{99.30(0)} & \textbf{100(0)} \\
				& \textbf{PUR} & 90.46(9.24) & 70.56(0.13) & 96.28(6.07) & 40.20(0) & \textbf{100(0)} & 69.50(0.39) & 60.00(0) & \underline{99.80(0)} & \textbf{100(0)} \\
				& \textbf{ARI} & 83.09(12.40) & 44.43(0.26) & 93.38(6.83) & 7.71(0) & \textbf{100(0)} & 51.04(0.81) & 49.26(0) & \underline{99.50(0)} & \textbf{100(0)} \\
				& \textbf{F-score} & 86.71(9.50) & 57.06(0.19) & 94.77(5.21) & 29.73(0) & \textbf{100(0)} & 61.38(0.69) & 62.03(0) & \underline{99.60(0)} & \textbf{100(0)} \\
				\hline
				\multirow{5}[2]{*}{\textbf{BBCSport}} & \textbf{ACC} & 78.27(5.06) & 81.19(0.17) & 82.21(0.36) & 49.82(0) & \underline{99.82(0)} & \underline{99.82(0)} & 72.43(0) & 98.90(0) & \textbf{100(0)} \\
				& \textbf{NMI} & 67.33(2.34) & 59.32(0.43) & 68.06(0.06) & 19.81(0) & \underline{99.38(0)} & \underline{99.38(0)} & 58.30(0) & 96.12(0) & \textbf{100(0)} \\
				& \textbf{PUR} & 81.78(2.23) & 81.19(0.17) & 82.46(0.18) & 50.00(0) & \underline{99.82(0)} & \underline{99.82(0)} & 73.53(0) & 98.90(0) & \textbf{100(0)} \\
				& \textbf{ARI} & 62.64(1.94) & 57.06(0.16) & 63.75(0.12) & 16.41(0) & \underline{99.77(0)} & \underline{99.77(0)} & 61.69(0) & 96.63(0) & \textbf{100(0)} \\
				& \textbf{F-score} & 70.95(1.48) & 68.34(0.16) & 71.84(0.10) & 41.32(0) & \underline{99.83(0)} & \underline{99.83(0)} & 72.20(0) & 97.43(0) & \textbf{100(0)} \\
				\hline
				\multirow{5}[2]{*}{\textbf{HW}} & \textbf{ACC} & 70.72(5.32) & 91.87(3.19) & 80.40(8.07) & 76.50(0) & 93.83(0.02) & 89.87(0.17) & 74.45(0) & \textbf{99.90(0)} & \underline{99.57(0.04)} \\
				& \textbf{NMI} & 68.48(2.24) & 86.29(1.24) & 77.59(4.52) & 73.27(0) & 89.26(0.07) & 81.27(0.36) & 73.52(0) & \textbf{99.76(0)} & \underline{98.92(0.10)} \\
				& \textbf{PUR} & 73.01(4.04) & 91.87(3.19) & 81.83(7.20) & 80.90(0) & 93.83(0.02) & 89.81(0.17) & 77.00(0) & \textbf{99.90(0)} & \underline{99.57(0.04)} \\
				& \textbf{ARI} & 58.30(4.39) & 83.48(2.31) & 71.16(7.88) & 65.95(0) & 88.51(0.09) & 79.10(0.36) & 53.31(0) & \textbf{99.78(0)} & \underline{99.06(0.09)} \\
				& \textbf{F-score} & 62.56(3.89) & 85.15(2.06) & 74.10(7.05) & 69.46(0) & 90.12(0.10) & 81.19(0.32) & 59.06(0) & \textbf{99.80(0)} & \underline{99.15(0.08)} \\
				\hline
				\multirow{5}[2]{*}{\textbf{Scene15}} & \textbf{ACC} & 50.51(2.60) & 54.01(0.62) & 52.39(2.80) & 60.11(0) & 77.86(0.44) & \underline{86.30(0.37)} & 23.55(0) & 73.97(0) & \textbf{88.64(1.82)} \\
				& \textbf{NMI} & 48.01(0.82) & 57.06(0.07) & 50.02(1.22) & 57.59(0) & 80.91(0.10) & \underline{87.84(0.48)} & 18.04(0) & 81.99(0) & \textbf{90.52(0.44)} \\
				& \textbf{PUR} & 54.40(1.73) & 61.67(0.90) & 57.36(2.44) & 62.30(0) & 83.58(0.21) & \underline{90.06(0.65)} & 25.08(0) & 74.55(0) & \textbf{90.19(1.18)} \\
				& \textbf{ARI} & 33.56(1.53) & 39.90(0.45) & 36.36(2.09) & 41.06(0) & 71.08(0.36) & \underline{82.03(0.87)} & 2.55(0) & 63.01(0) & \textbf{84.91(1.31)} \\
				& \textbf{F-score} & 38.13(1.40) & 44.62(0.43) & 40.75(1.94) & 45.14(0) & 73.17(0.34) & \underline{83.27(0.81)} & 14.77(0) & 70.09(0) & \textbf{85.99(1.21)} \\
				\hline
				\multirow{5}[2]{*}{\textbf{MITIndoor}} & \textbf{ACC} & 42.98(1.58) & 38.18(1.95) & 45.06(1.21) & 20.35(0) & 68.41(0.52)    & \underline{74.65(1.02)}    & 42.51(0)    & 53.06(0) & \textbf{82.54(1.90)} \\
				& \textbf{NMI} & 56.39(0.79) & 52.30(0.95) & 57.55(0.83) & 32.31(0) & 74.82(0.73)    & \underline{87.96(0.24)}    & 59.39(0)    & 53.72(0) & \textbf{92.96(0.54)} \\
				& \textbf{PUR} & 44.82(1.39) & 40.36(1.66) & 47.17(0.98) & 22.16(0) & 69.36(0.48)    & \underline{76.59(0.17)}    & 46.25(0)    & 64.91(0) & \textbf{85.63(1.62)} \\
				& \textbf{ARI} & 29.76(1.01) & 22.94(1.25) & 31.20(1.29) & 4.94(0) & 55.49(0.51)    & \underline{70.62(0.10)}    & 37.70(0)    & 44.75(0) & \textbf{79.25(2.24)} \\
				& \textbf{F-score} & 30.83(0.99) & 24.23(1.21) & 32.25(1.27) & 7.11(0) & 56.17(0.76)    & \underline{71.54(0.18)}    & 39.85(0)    & 49.50(0) & \textbf{79.58(2.20)} \\
				\hline
				\multirow{5}[2]{*}{\textbf{CCV}} & \textbf{ACC} & 18.30(0.82) & 21.99(0.75) & 22.00(0.22) & 20.74(0) & \underline{42.95(0.06)}    & 26.20(0.06)    & 17.95(0)    & 22.09(0)    & \textbf{72.01(1.62)} \\
				& \textbf{NMI} & 14.45(0.51) & 18.91(0.42) & 18.05(0.27) & 17.34(0) & \underline{39.86(0.05)}    & 20.12(0.69)    & 16.54(0)    & 15.61(0)    & \textbf{75.92(0.74)} \\
				& \textbf{PUR} & 21.86(0.55) & 25.23(0.47) & 25.38(0.23) & 22.84(0) & \underline{32.57(0.04)}    & 26.85(0.71)    & 22.89(0)    & 24.60(0)    & \textbf{74.35(1.14)} \\
				& \textbf{ARI} & 5.82(0.47) & 7.17(0.23) & 7.82(0.11) & 6.23(0) & \underline{22.63(0.04)}   & 14.42(0.41)    & 5.95(0)    & 10.81(0)    & \textbf{62.52(2.15)} \\
				& \textbf{F-score} & 10.96(0.44) & 12.96(0.21) & 12.82(0.11) & 13.25(0) & \underline{31.65(0.05)}    & 23.01(0.34)    & 11.08(0)    & 20.50(0)    & \textbf{64.71(2.02)} \\
				\hline
				\multirow{5}[2]{*}{\textbf{Caltech101-all}} & \textbf{ACC} & 24.89(1.37) & 26.90(1.05) & \underline{27.19(1.37)} & 20.94(0) & OM    & OM    & OM    & OM    & \textbf{49.04(1.43)} \\
				& \textbf{NMI} & 48.61(0.68) & 48.60(0.48) & \underline{49.61(0.52)} & 35.76(0) & OM    & OM    & OM    & OM    & \textbf{84.10(0.35)} \\
				& \textbf{PUR} & 46.53(0.95) & 46.46(0.50) & \underline{48.09(0.62)} & 34.18(0) & OM    & OM    & OM    & OM    & \textbf{75.61(0.85)} \\
				& \textbf{ARI} & 19.41(1.58) & \underline{22.80(1.27)} & 20.80(1.07) & 5.91(0) & OM    & OM    & OM    & OM    & \textbf{35.76(1.03)} \\
				& \textbf{F-score} & 20.66(1.57) & \underline{24.28(1.26)} & 22.03(1.07) & 9.82(0) & OM    & OM    & OM    & OM    & \textbf{36.74(1.02)} \\
				\hline
				\multirow{5}[2]{*}{\textbf{Animal}} & \textbf{ACC} & 15.20(0.77) & 17.30(0.22) & 17.22(0.10) & \underline{18.59(0)} & OM    & OM    & OM    & OM    & \textbf{88.61(2.11)} \\
				& \textbf{NMI} & 11.79(0.51) & 12.38(0.08) & \underline{14.11(0.19)} & 13.04(0) & OM    & OM    & OM    & OM    & \textbf{90.47(1.05)} \\
				& \textbf{PUR} & 18.21(0.53) & 19.41(0.14) & 18.44(0.20) & \underline{20.40(0)} & OM    & OM    & OM    & OM    & \textbf{90.49(1.84)} \\
				& \textbf{ARI} & 4.29(0.36) & 4.61(0.04) & 6.33(0.04) & \underline{6.99(0)} & OM    & OM    & OM    & OM    & \textbf{84.85(2.19)} \\
				& \textbf{F-score} & 9.52(0.35) & 10.61(0.08) & \underline{14.21(0.02)} & 13.00(0) & OM    & OM    & OM    & OM    & \textbf{85.71(2.05)} \\
				\hline
				\multirow{5}[2]{*}{\textbf{NUSWIDEOBJ}} & \textbf{ACC} & 14.17(0.49) & 13.11(0.24) & 16.85(0.17) & \underline{17.49(0)} & OM    & OM    & OM    & OM    & \textbf{18.94(0.39)} \\
				& \textbf{NMI} & \underline{12.59(0.30)} & 12.06(0.22) & 12.25(0.08) & 12.57(0) & OM    & OM    & OM    & OM    & \textbf{23.09(0.06)} \\
				& \textbf{PUR} & 23.62(0.56) & 23.65(0.39) & 22.73(0.29) & \underline{24.61(0)} & OM    & OM    & OM    & OM    & \textbf{31.94(0.33)} \\
				& \textbf{ARI} & 4.74(0.27) & 3.88(0.13) & \underline{5.65(0.26)} & 4.80(0) & OM    & OM    & OM    & OM    & \textbf{7.93(0.11)} \\
				& \textbf{F-score} & 8.83(0.26) & 8.42(0.13) & 11.11(0.20) & \underline{11.69(0)} & OM    & OM    & OM    & OM    & \textbf{12.09(0.11)} \\
				\hline
			\end{tabular}%
		}
		
		\label{tab:results}%
	\end{table*}%
	
\subsection{Experimental Results}
Table~\ref{tab:results} provides a comprehensive overview of the clustering results for all methods across 9 datasets. The best and second-best results are highlighted using \textbf{bold} and \underline{underlined} values, respectively. "OM" denotes unavailable results due to out-of-memory errors. Based on the results, we can draw the following conclusions:

(1) DSTL outperforms the baselines across most datasets, with particularly notable improvements observed on the CCV and Animal datasets. For example, on the CCV dataset, DSTL achieves enhancements of 29.06\%, 36.06\%, 41.78\%, 39.89\%, and 33.06\% over the second-best method t-SVD-MSC across five evaluation metrics. It can be attributed to the presence of considerable semantic-unrelated information in these datasets, which adversely affects the clustering performance of the baseline methods. However, DSTL effectively disentangles the latent features, mitigating the negative impact of such irrelevant information and resulting in superior clustering performance, demonstrating the efficiency and advantages of our feature disentanglement strategy.

(2) Both our proposed DSTL and other tensor-based methods (i.e., t-SVD-MSC, ETLMSC, TBGL and LTBPL) consistently outperform non-tensor approaches (i.e., AWMVC, FDAGF, FSMSC, and UDBGL) on the first six datasets. This is primarily due to the better capacity of tensor regulaization in leveraging complementary information and the capturing high-order correlations among multiple views, which non-tensor methods lack.

(3) The DSTL outperforms the majority of the other tensor-based methods on the first six datasets. This is attributed to its unique ability to consider both disentangled semantically unrelated and related features. By integrating a consensus alignment indicator into the disentanglement process, DSTL optimizes the clustering outcomes by ensuring that the extracted features are aligned across multiple views, leading to superior performance.

(4) Additionally, compared to other tensor-based methods that stack affinity graphs as tensors, DSTL offers a significant advantage in terms of memory efficiency. Traditional tensor stacking approaches often lead to high space complexity of $O(n^2)$, resulting in memory errors on large-scale datasets. However, DSTL directly operates on latent low-dimensional features, which have a linear space complexity of $O(n)$, making it more memory-efficient and scalable to larger datasets.

\subsection{Parameter Analysis}
\begin{figure*}[!t]
	\centering
	\subfigure[NGs]{\includegraphics[width=4.25cm,height=3.2cm]{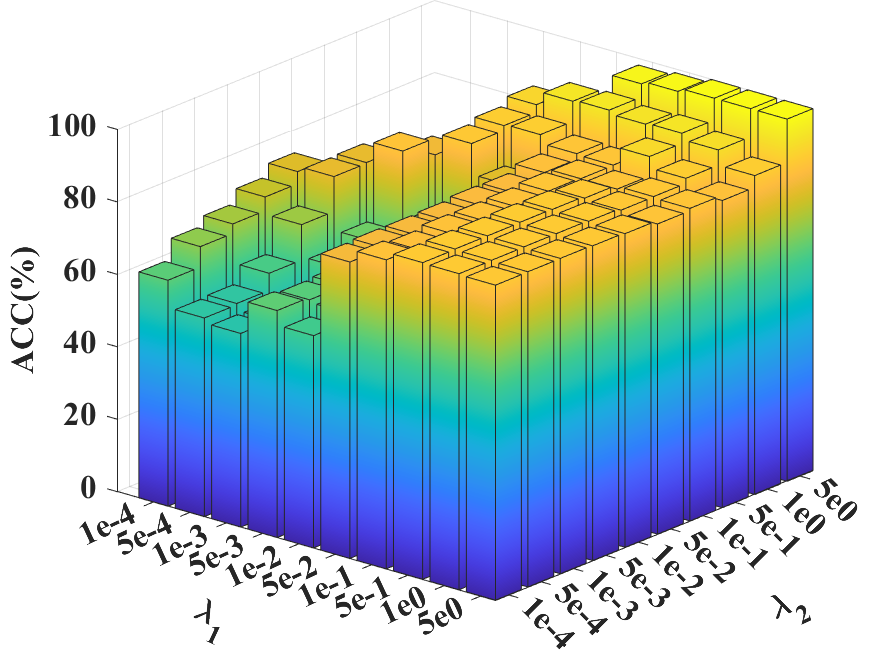}}
	\subfigure[BBCSport]{\includegraphics[width=4.25cm,height=3.2cm]{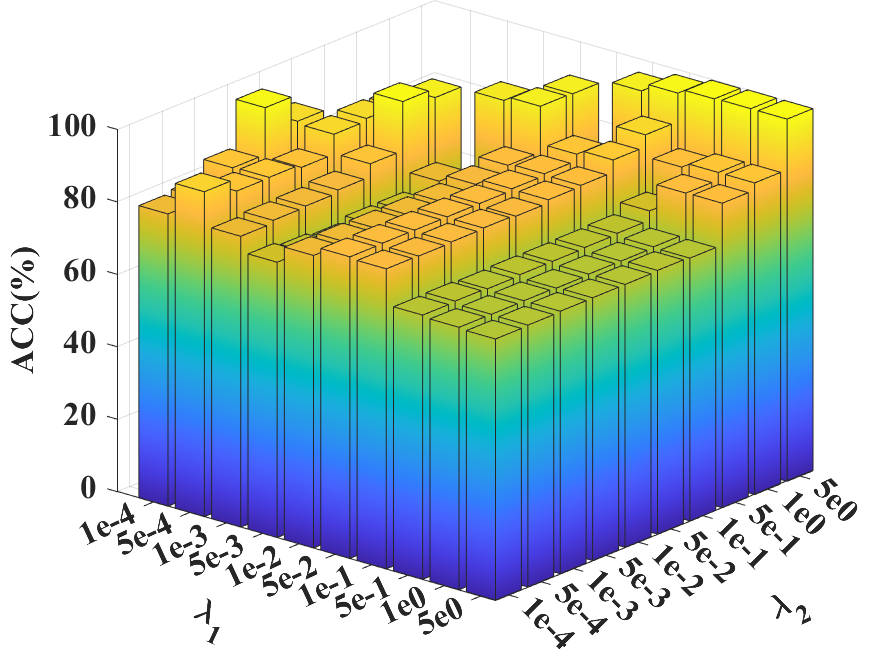}}
	\subfigure[HW]{\includegraphics[width=4.25cm,height=3.2cm]{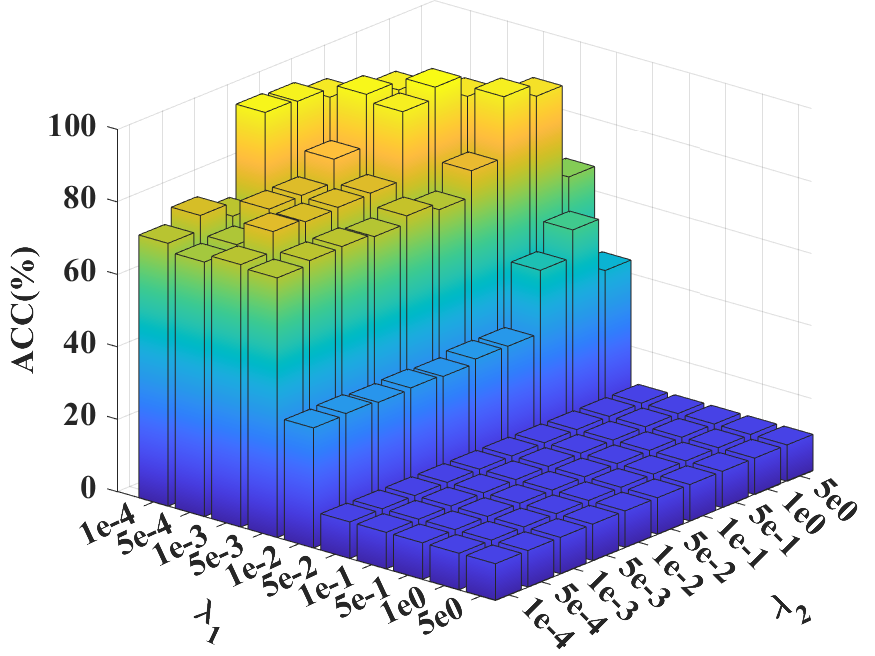}}
	\subfigure[NUSWIDEOBJ]{\includegraphics[width=4.25cm,height=3.2cm]{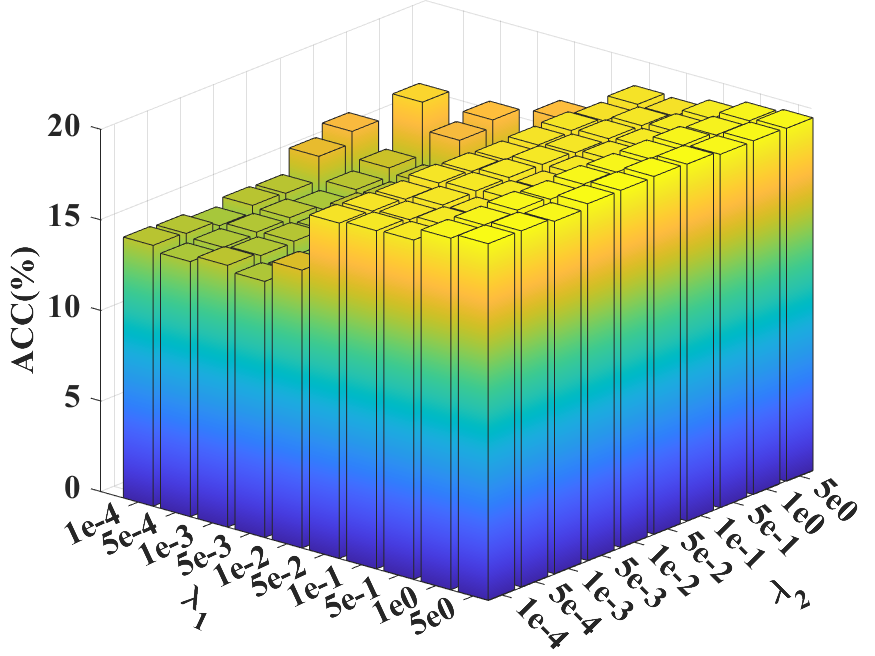}}
	\caption{The sensitivity analysis of clustering results of DSTL w.r.t. $\lambda_1$ and $\lambda_2$ on (a) NGs, (b) BBCSport, (c) HW and (d) NUSWIDEOBJ datasets.}
	\label{parameter12}
\end{figure*}

In this section, we investigate the impact of different parameters in DSTL (i.e., $\lambda_1$, $\lambda_2$, $\lambda_3$, and $k$) on clustering performance and examine their interrelationship across the NGs, BBCSport, HW, and NUSWIDEOBJ datasets. Specifically, we first fix $\lambda_3$ and $k$, then analyze the influence of $\lambda_1$ and $\lambda_2$ on the ACC. The results are depicted in Fig.~\ref{parameter12}. Notably, we observe that DSTL exhibits higher sensitivity to variations in $\lambda_1$ compared to $\lambda_2$. This phenomenon can be attributed to the fact that $\lambda_1$ encodes the regularization of the negative influence of disentangled semantic-unrelated features. However, DSTL consistently achieves satisfactory results across a relatively broad range for both parameters. 

\begin{figure*}[!t]
	\centering
	\subfigure[NGs]{\includegraphics[width=4.25cm,height=3.2cm]{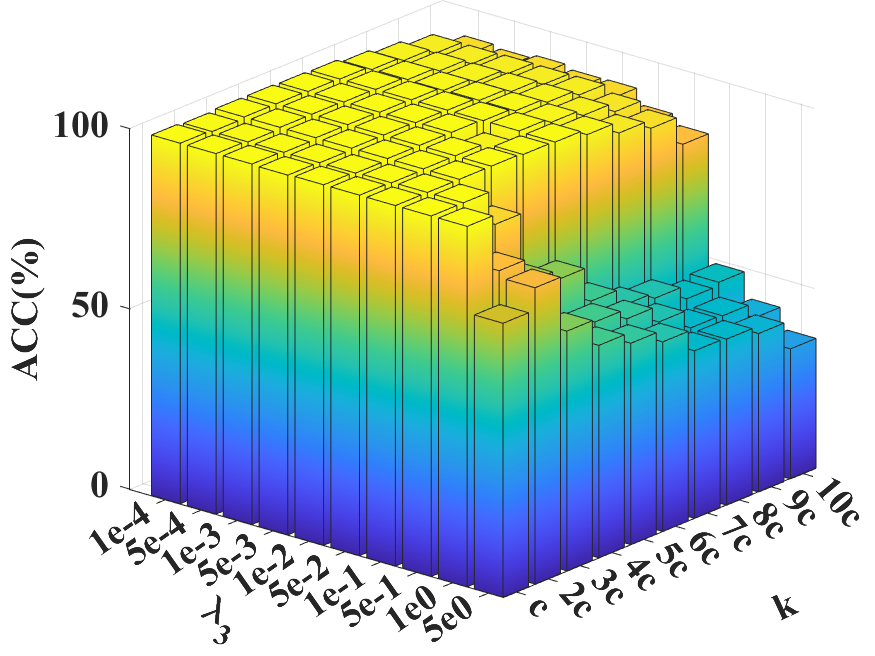}}
	\subfigure[BBCSport]{\includegraphics[width=4.25cm,height=3.2cm]{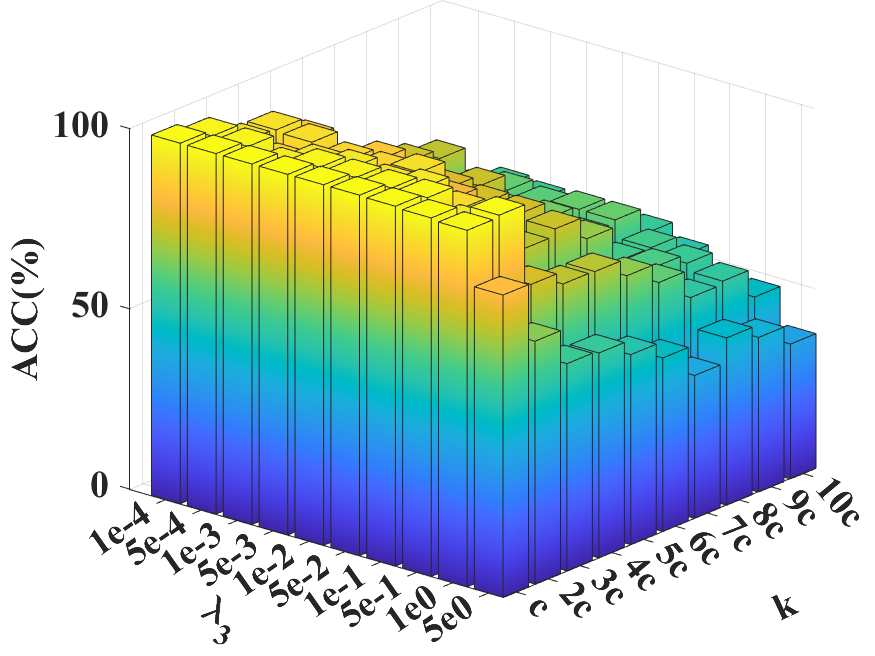}}
	\subfigure[HW]{\includegraphics[width=4.25cm,height=3.2cm]{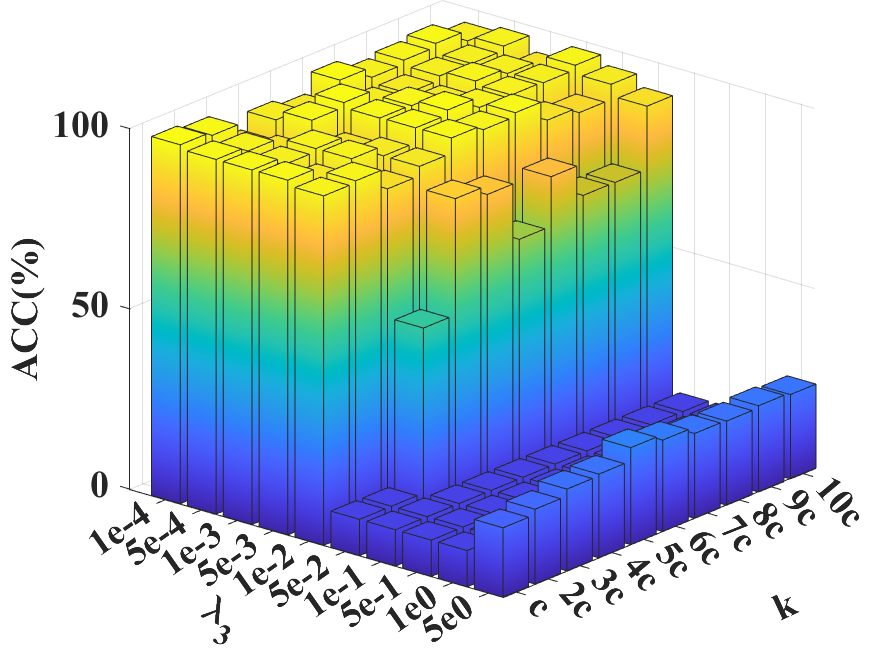}}
	\subfigure[NUSWIDEOBJ]{\includegraphics[width=4.25cm,height=3.2cm]{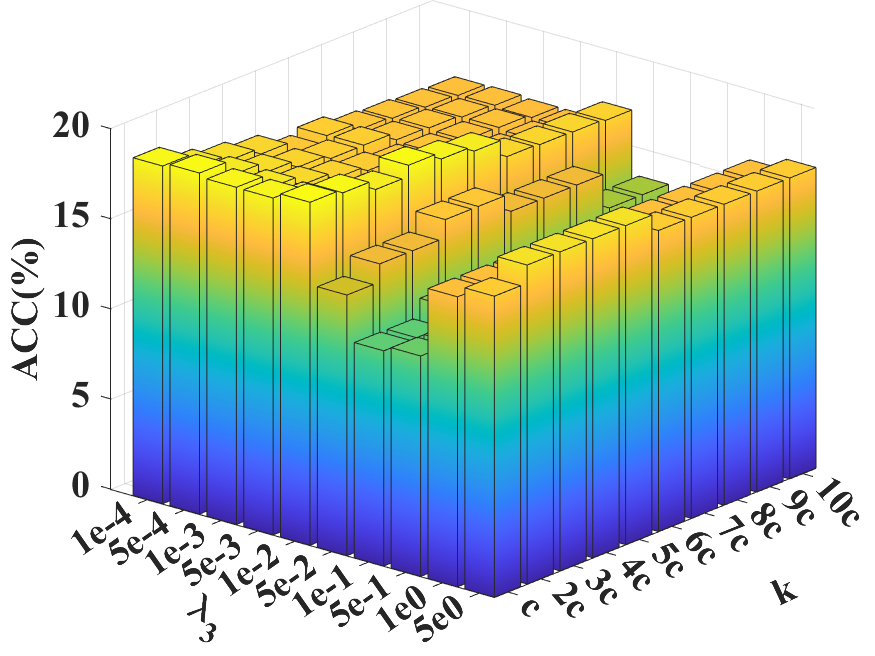}}
	\caption{The sensitivity analysis of clustering results of DSTL w.r.t. $\lambda_3$ and $k$ on (a) NGs, (b) BBCSport, (c) HW and (d) NUSWIDEOBJ datasets.}
	\label{parameter3}
\end{figure*}

When analyzing $\lambda_3$ and $k$, we keep $\lambda_1$ and $\lambda_2$ fixed. Fig.~\ref{parameter3} presents the outcomes of ACC value w.r.t. various combinations of $\lambda_3$ and $k$. Remarkably, when $\lambda_3$ is in the range of $[1e-4,1e-2]$, DSTL consistently achieves excellent performance across the four datasets. This indicates that DSTL has low sensitivity to changes in $\lambda_3$, reflecting the effectiveness of the proposed consensus semantic alignment strategy. For the latent representation dimension $k$, we find that setting $k=c$ results in the best or satisfactory ACC values across the four datasets. Therefore, we simply fixing $\lambda_3=1e-4$ and $k=c$ for all datasets to yield satisfactory clustering results and avoid parameter search. 

\begin{figure*}[!t]
	\centering
	\subfigure[NGs]{\includegraphics[width=4.3cm,height=3cm]{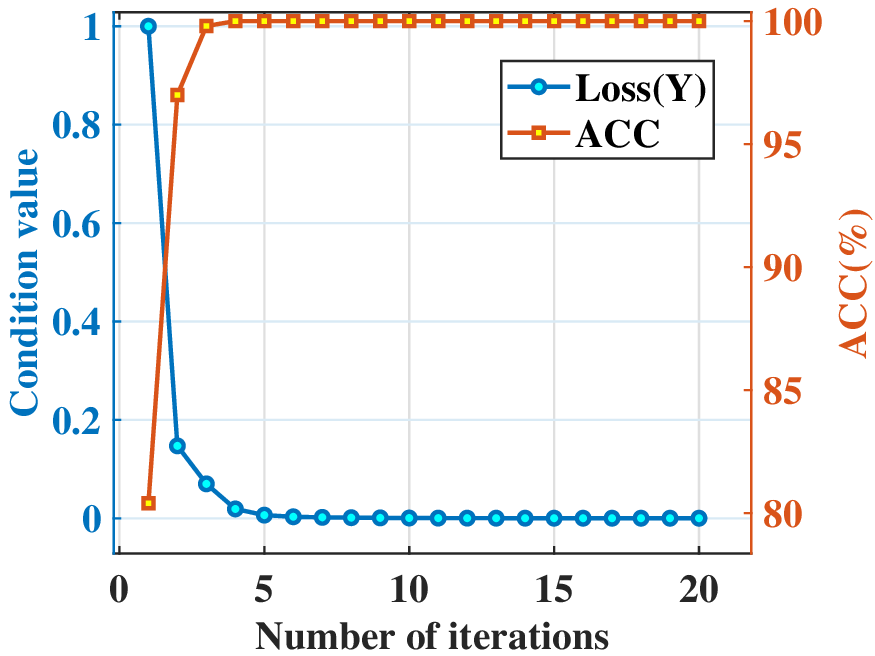}}
	\subfigure[BBCSport]{\includegraphics[width=4.3cm,height=3cm]{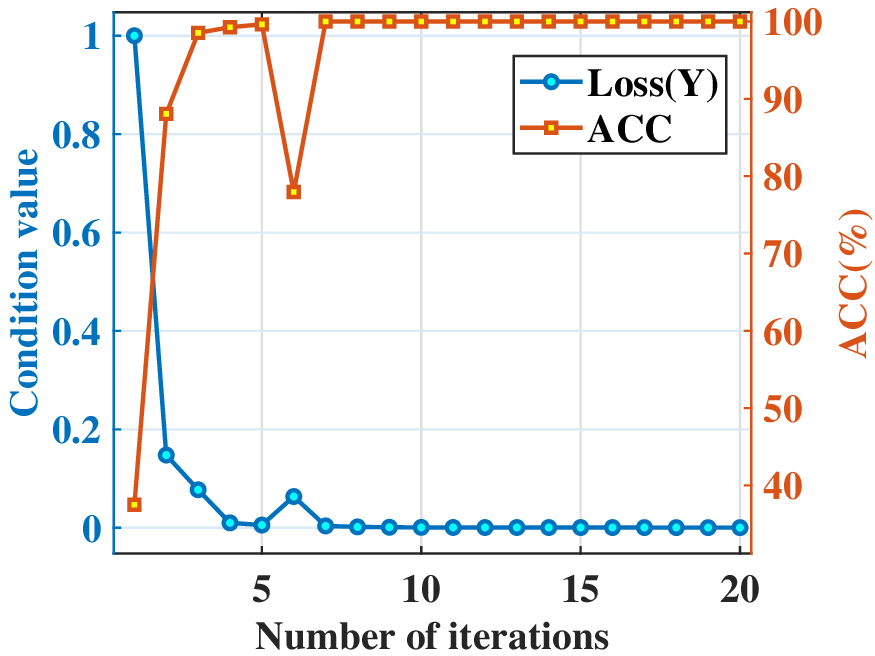}}
	\subfigure[HW]{\includegraphics[width=4.3cm,height=3cm]{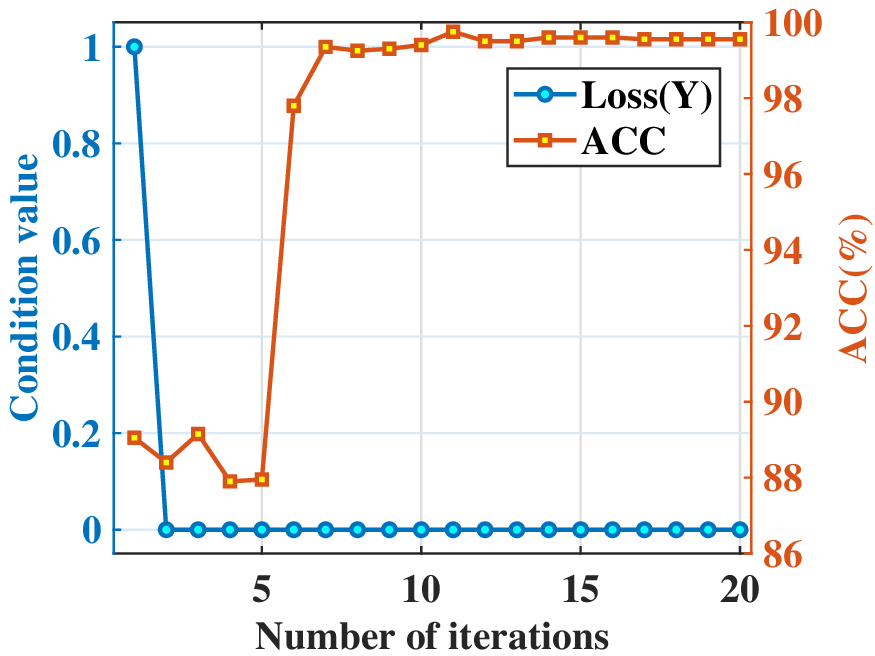}}
	\subfigure[NUSWIDEOBJ]{\includegraphics[width=4.3cm,height=3cm]{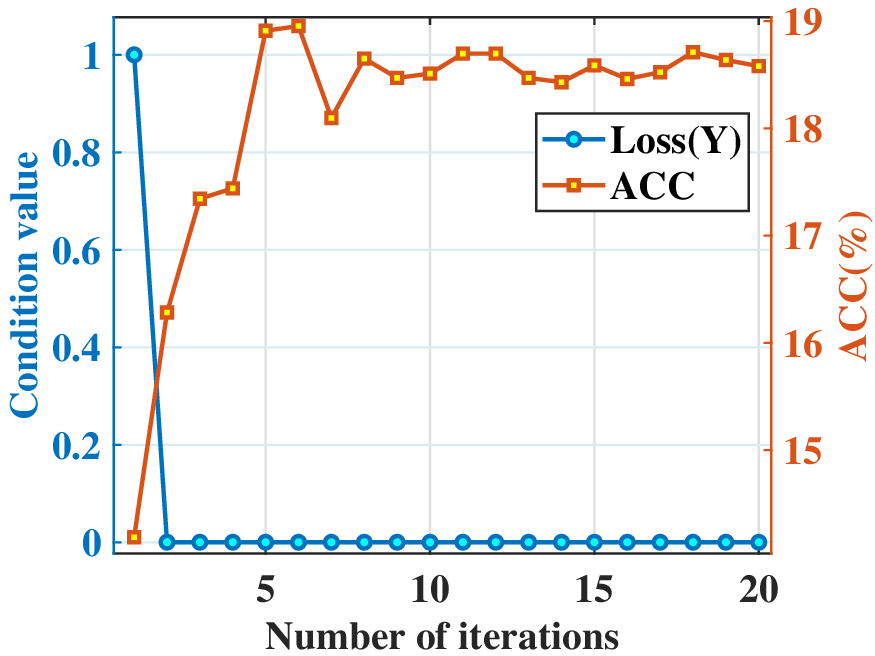}}
	\caption{The condition value and ACC w.r.t. the number of iterations on (a) NGs, (b) BBCSport, (c) HW and (d) NUSWIDEOBJ datasets.}
	\label{converge}
\end{figure*}

\begin{table*}[!t]
	\renewcommand{\arraystretch}{1.15}
	\centering
	\caption{Running time (second) of baselines and the proposed DSTL on nine datasets. OM indicates \textit{out of memory}.}
		\resizebox{\linewidth}{!}{
			\begin{tabular}{c|ccccccccc}
				\hline
				\textbf{Dataset} & \textbf{AWMVC} & \textbf{FDAGF} & \textbf{FSMSC} & \textbf{UDBGL} & \textbf{t-SVD-MSC} & \textbf{ETLMSC} & \textbf{TBGL} & \textbf{LTBPL} & \textbf{Ours} \\
				\hline
				\textbf{NGs} & 1.71  & 1.59  & 3.52  & 1.58  & 10.95 & \underline{0.81}  & 10.05 & 4.36  & \textbf{0.13} \\
				\textbf{BBCSport} & 2.41  & 1.78  & 3.66  & \underline{0.56}  & 8.67  & 0.76  & 9.34  & 4.78  & \textbf{0.09} \\
				\textbf{HW} & \underline{1.03}  & 7.89  & 10.46 & 24.31 & 296.45 & 28.97 & 492.31 & 155.47 & \textbf{0.41} \\
				\textbf{Scene15} & \underline{12.34} & 41.49 & 31.74 & 32.72 & 487.27 & 60.98 & 2187.56 & 1378.48 & \textbf{1.35} \\
				\textbf{MITIndoor} & \underline{63.03} & 170.98 & 72.14 & 154.31 & 3896.39    & 625.26    & 5222.61    & 2346.98 & \textbf{6.19} \\
				\textbf{CCV} & \textbf{0.54} & 13.06 & 36.91 & 67.49 & 7283.52  & 1095.64    & 9102.97    & 6803.31    & \underline{1.65} \\
				\textbf{Caltech101-all} & \underline{34.65} & 129.34 & 112.72 & 322.07 & OM    & OM    & OM    & OM    & \textbf{12.94} \\
				\textbf{Animal} & \underline{64.88} & 351.32 & 115.17 & 162.08 & OM    & OM    & OM    & OM    & \textbf{5.75} \\
				\textbf{NUSWIDEOBJ} & \underline{28.68} & 303.03 & 225.98 & 463.22 & OM    & OM    & OM    & OM    & \textbf{13.79} \\
				\hline
			\end{tabular}%
		}
		\label{time}%
	\end{table*}%
		
		\begin{figure*}[t]
			\centering
			\subfigure[BBCSport]{\includegraphics[width=4.35cm,height=3.4cm]{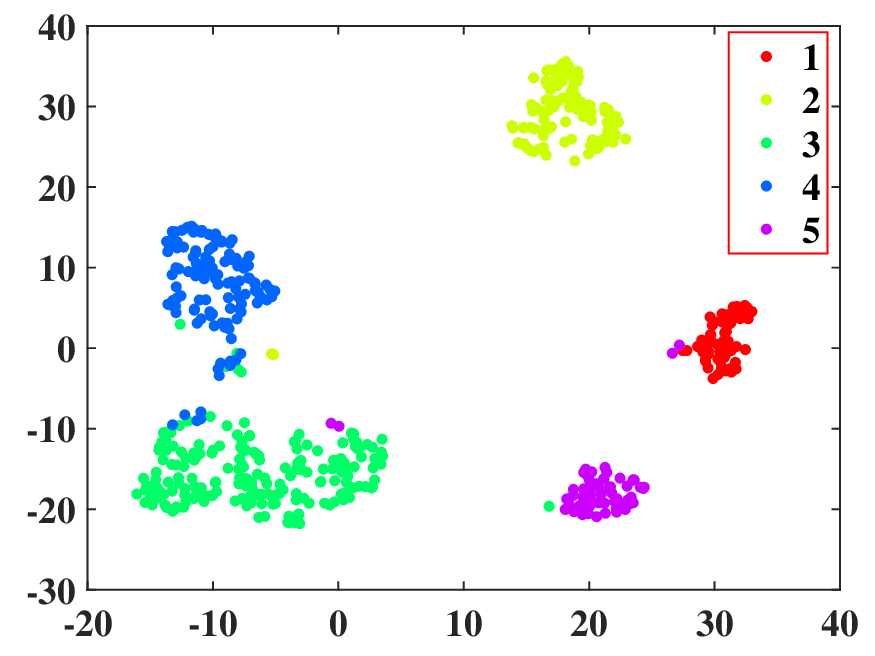}\includegraphics[width=4.35cm,height=3.4cm]{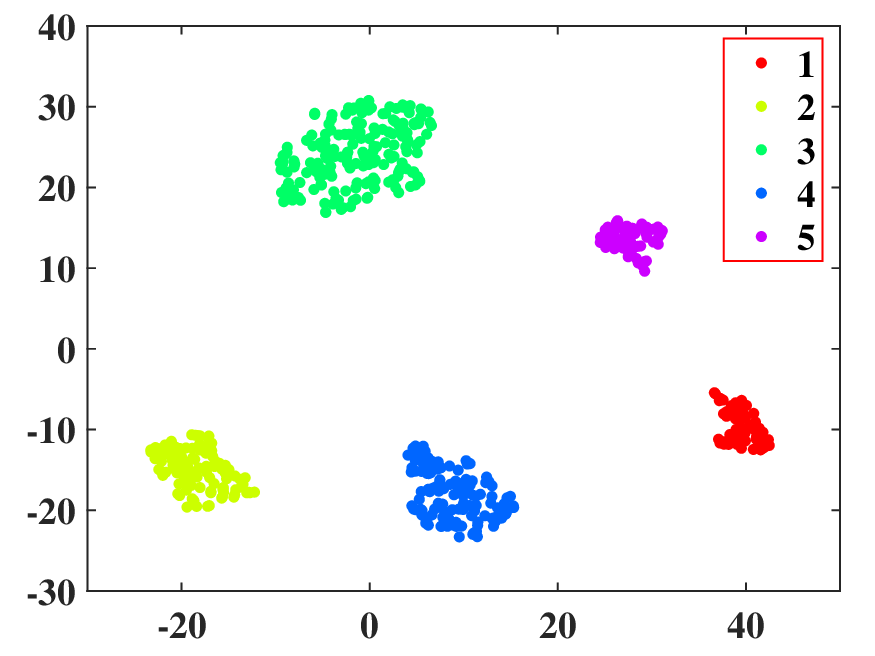}}
			\subfigure[HW]{\includegraphics[width=4.35cm,height=3.4cm]{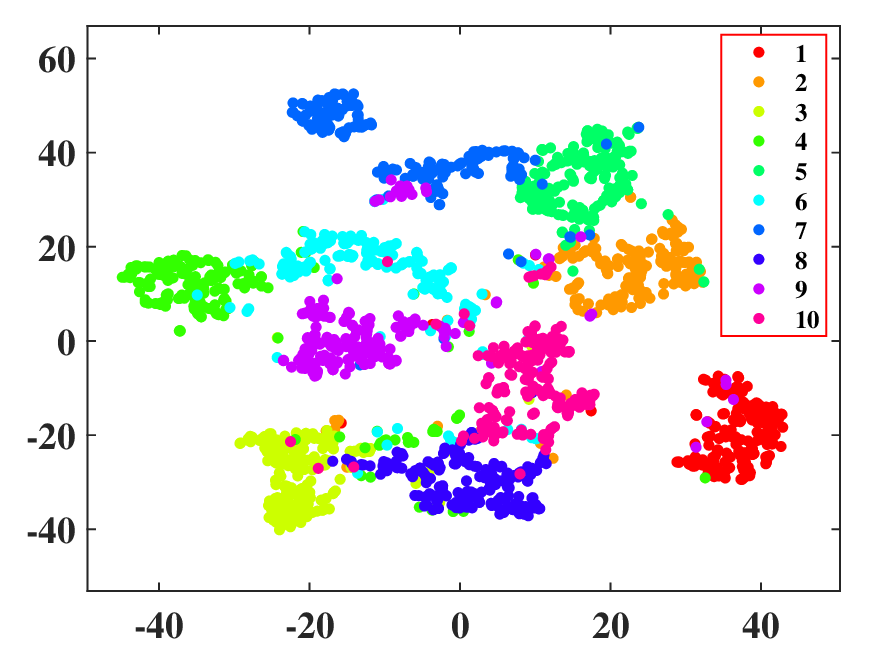}\includegraphics[width=4.35cm,height=3.4cm]{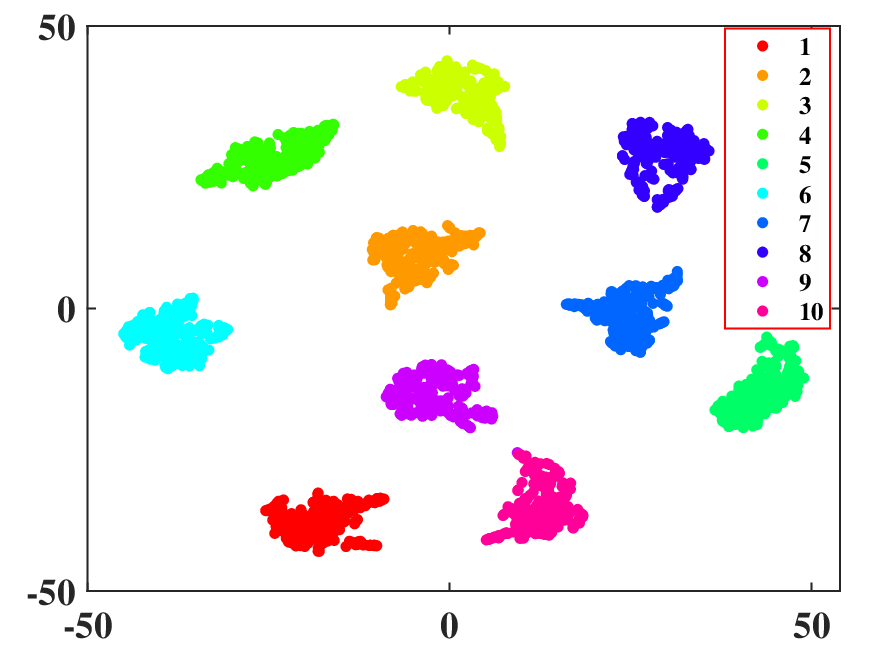}}\\
			\subfigure[Scene15]{\includegraphics[width=4.35cm,height=3.4cm]{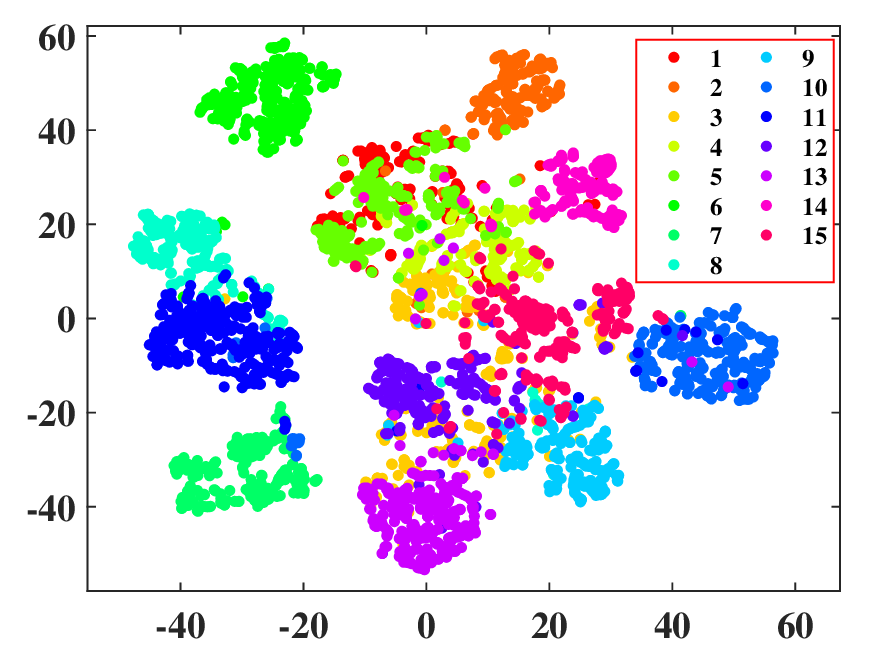}\includegraphics[width=4.35cm,height=3.4cm]{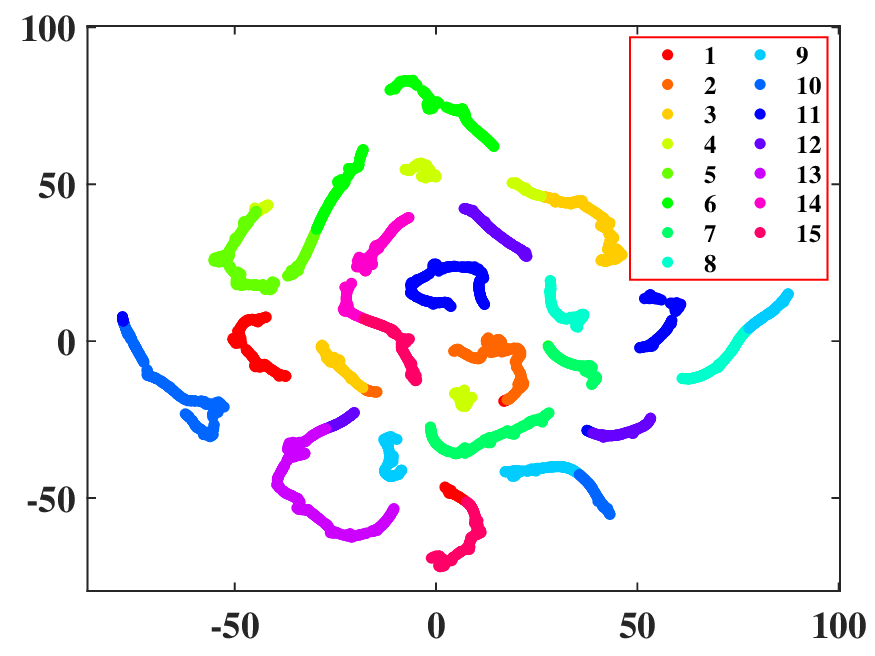}}
			\subfigure[Animal]{\includegraphics[width=4.35cm,height=3.4cm]{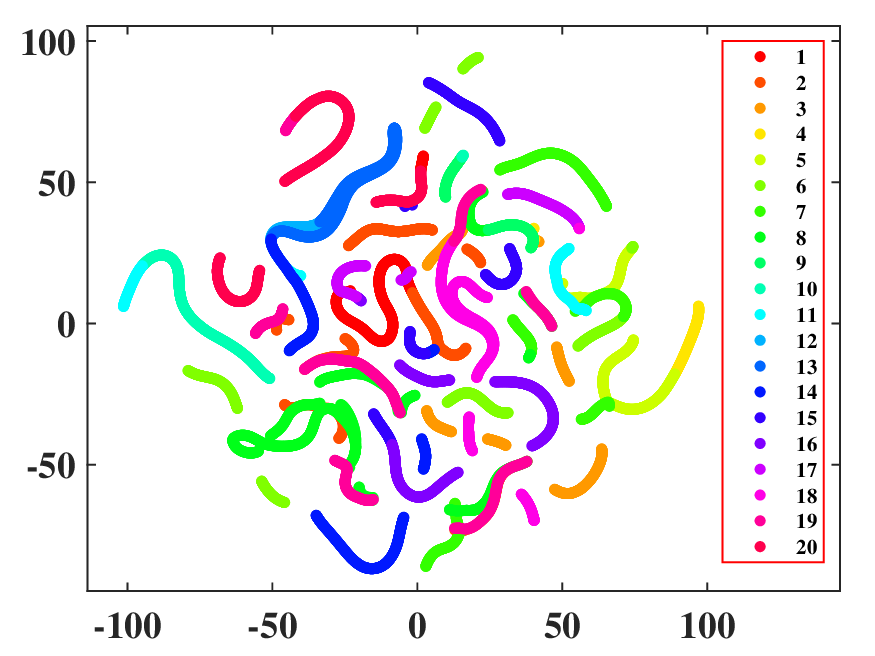}\includegraphics[width=4.35cm,height=3.4cm]{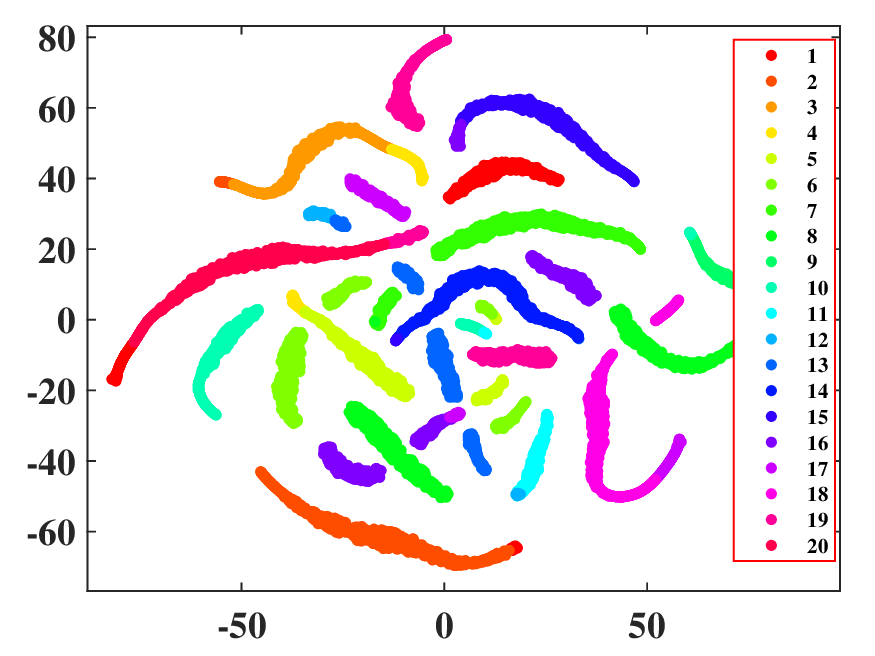}}
			\caption{The t-SNE visualizations of learned consensus alignment indicator obtained by DSTL-S (left) and DSTL (right) on (a) BBCSport, (b) HW, (c) Scene15 and (d) Animal datasets.}
			\label{visualization}
		\end{figure*}


\subsection{Convergence and Time Comparison}
\subsubsection{Convergence Analysis}
As illustrated before, the objective value of DSTL decreases monotonically with variables alternate update and
the objective function is lower-bounded. In this subsection, we experimentally demonstrate the convergence property of the optimization algorithm for Algorithm 1. We set the convergence condition as $Loss(\mathbf{Y})=\|\mathbf{Y}_{t}-\mathbf{Y}_{t-1}\|_F^2/\|\mathbf{Y}_{t-1}\|_F^2\le 1e-4$. Fig.~\ref{converge} displays the change in $Loss(Y)$ and ACC values w.r.t. iterations on NGs, BBCSport, HW, and NUSWIDEOBJ datasets. It is noticeable that $Loss(\mathbf{Y})$ experiences rapid reduction and typically converges to a stable value within 15-20 iterations. Meanwhile, the ACC value also exhibits swift increase and stabilizes promptly. The outcomes demonstrate the favorable convergence property inherent in our algorithm.
\subsubsection{Time Comparison}
Table~\ref{time} reports the running time of all methods on all benchmark datasets. It is evident that our DSTL achieves competitive computational efficiency compared to almost all baselines and notably outperforms other tensor-based methods. This efficiency is attributed to our slim tensor learning strategy, which effectively reduces the dimensions of the representation tensor. In contrast, other tensor-based methods suffer from extremely high time complexity (i.e., $O(n^3)$ or $O(n^2log(n))$), rendering them unsuitable for large-scale datasets. While some anchor graph-based approaches (e.g., FDAGF, FSMSC, and UDBGL) exhibit $O(n)$ time complexity, they need to solve a QP problem w.r.t. each sample to seek the sample-anchor similarities, which can reduce their efficiency in practice. However, all variables of DSTL are updated directly with closed-form solutions, requiring only 15-20 iterations to achieve optimal results. Thus, our method generally spends less time than those baselines. These results demonstrate that our DSTL possesses greater potential for handling large-scale MVC tasks in practical application.

\begin{figure*}[!t]
	\centering
	\subfigure[NGs]{\includegraphics[width=5.55cm,height=3.2cm]{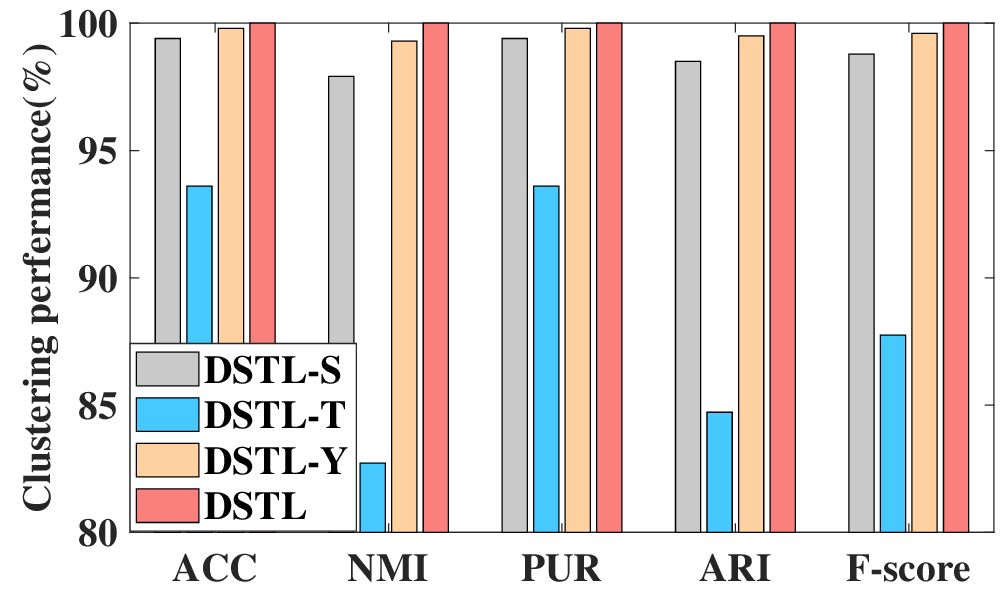}}
	\hspace{0.1cm}
	\subfigure[BBCSport]{\includegraphics[width=5.55cm,height=3.2cm]{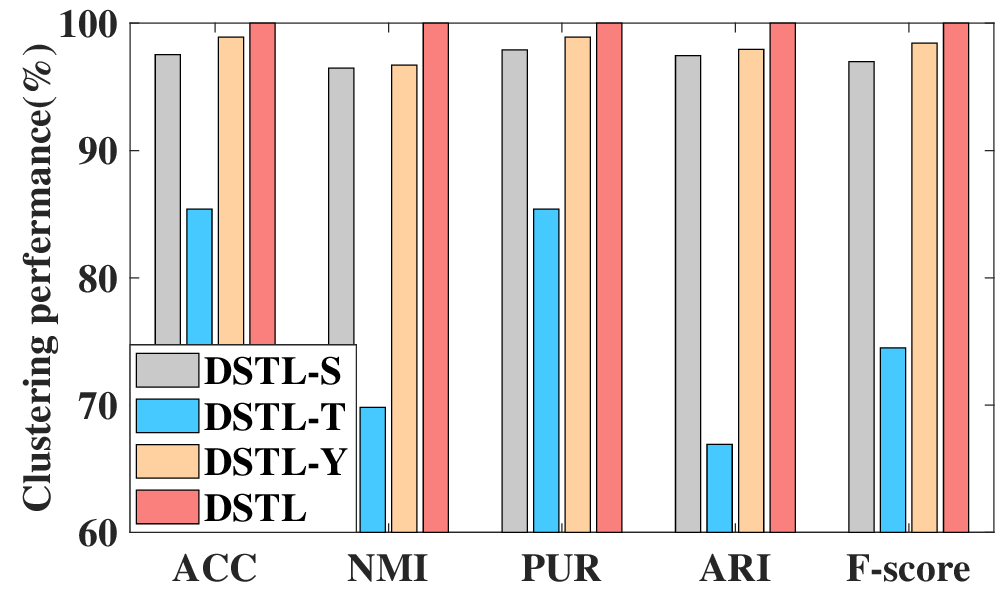}}
	\hspace{0.1cm}
	\subfigure[HW]{\includegraphics[width=5.55cm,height=3.2cm]{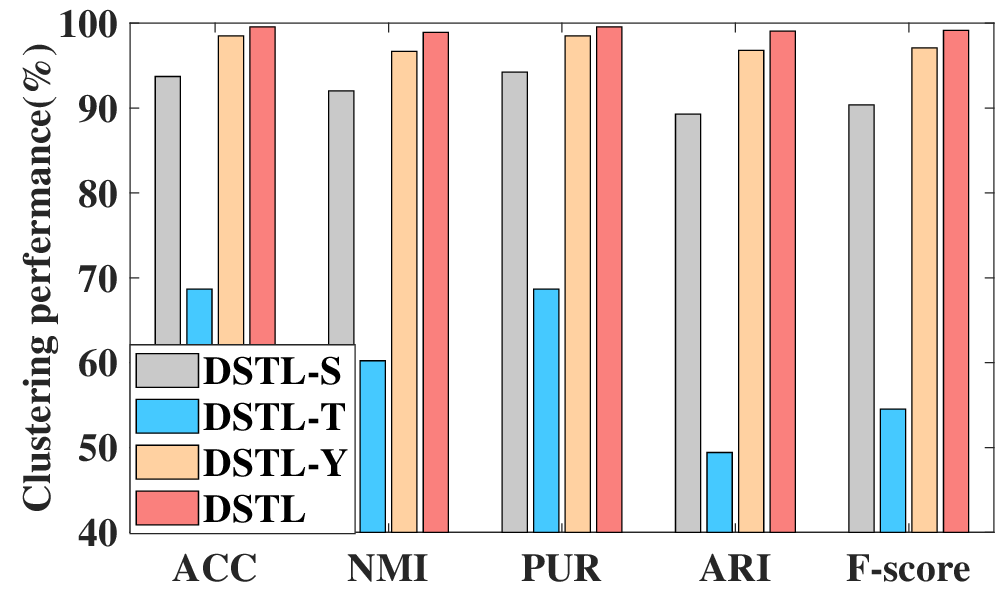}}\\
	\subfigure[Scene15]{\includegraphics[width=5.55cm,height=3.2cm]{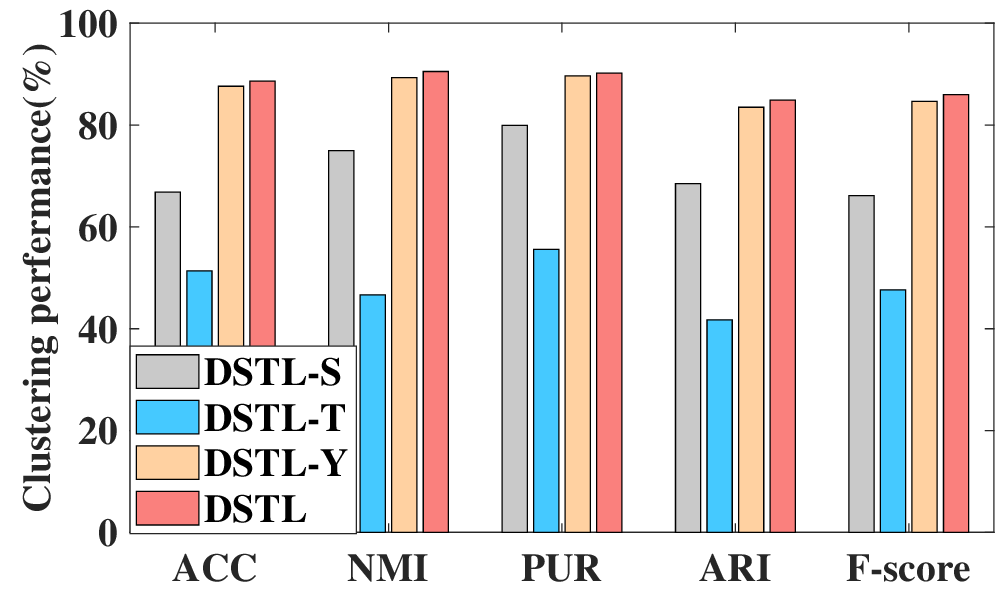}}
	\hspace{0.1cm}
	\subfigure[MITIndoor]{\includegraphics[width=5.55cm,height=3.2cm]{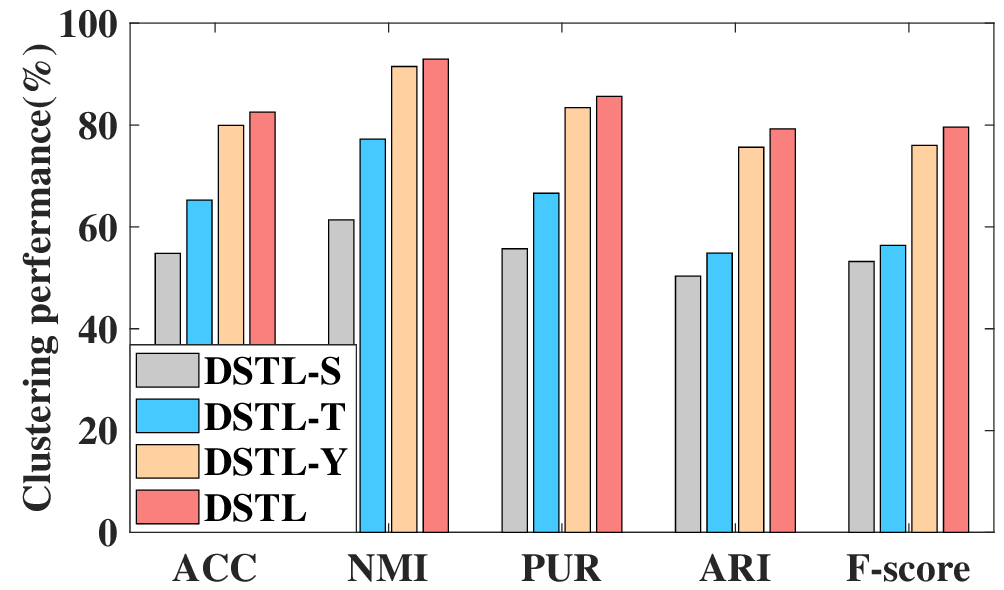}}
	\hspace{0.1cm}
	\subfigure[CCV]{\includegraphics[width=5.55cm,height=3.2cm]{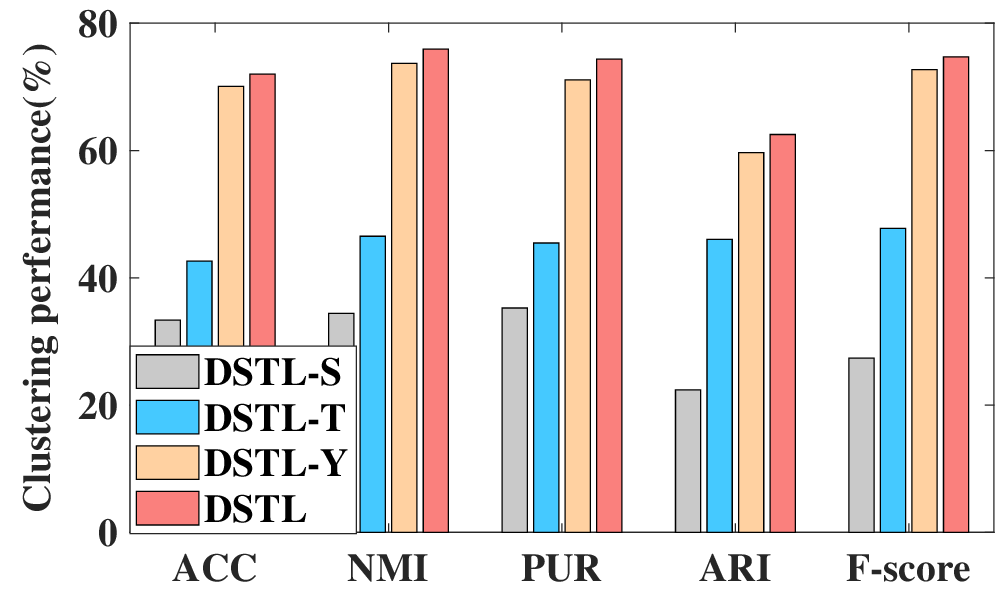}}\\
	\subfigure[Caltech101-all]{\includegraphics[width=5.55cm,height=3.2cm]{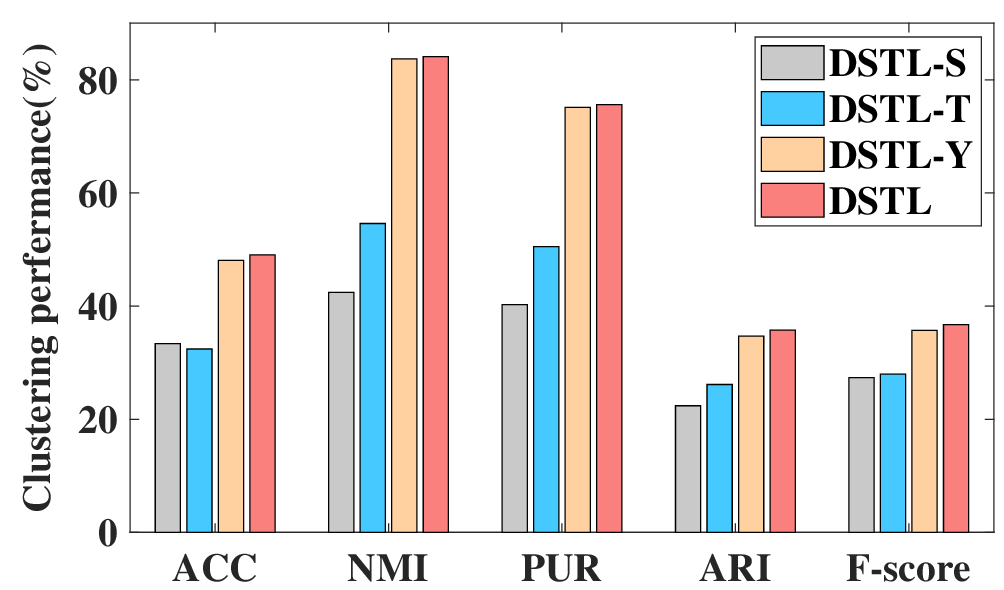}}
	\hspace{0.1cm}
	\subfigure[Animal]{\includegraphics[width=5.55cm,height=3.2cm]{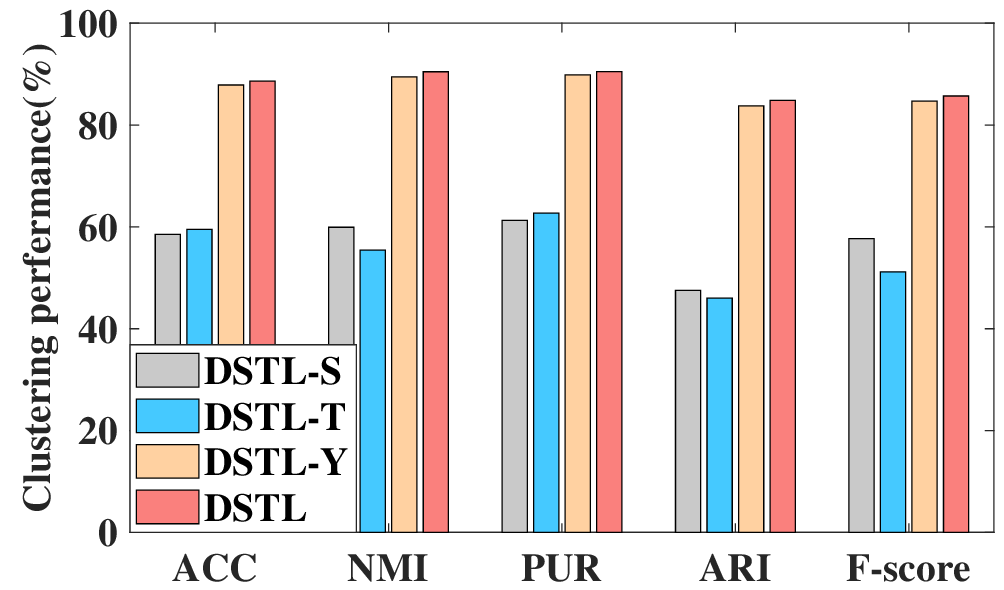}}
	\hspace{0.1cm}
	\subfigure[NUSWIDEOBJ]{\includegraphics[width=5.55cm,height=3.2cm]{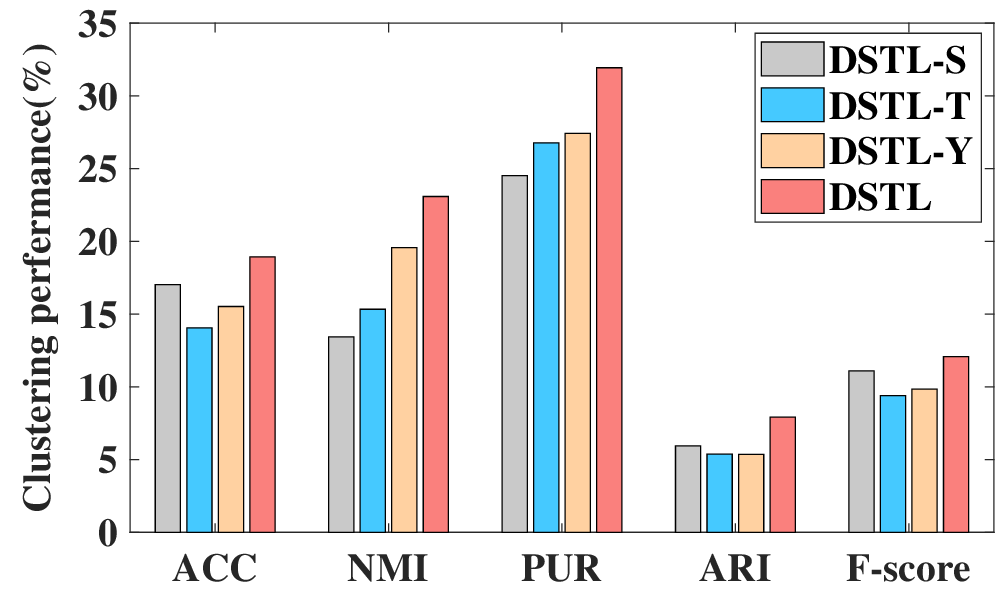}}
	\caption{The ablation study results on nine datasets.}
	\label{abl}
\end{figure*}

\subsection{Ablation Study}
To assess the effectiveness of multi-view features disentanglement and tensor-based regularization, as well as the consensus semantic alignment indicator learning, we derive three variants: DSTL-S, DSTL-T, and DSTL-Y.  

(1) DSTL-S drops the semantic-unrelated features learning and just focus on the semantic-related representations. The loss function of DSTL-S is

\begin{equation}\label{AS}\small
    \begin{aligned}
	\min_{\Omega} &\sum_{v=1}^{m}\|\mathbf{X}^{v}-\mathbf{W}^{v}\mathbf{H}^{v}\|_F^2+\lambda_2\|\mathcal{H}\|_{\circledast}+\lambda_3\sum_{v=1}^{m}\|\mathbf{H}^{v}-\mathbf{C}^{v}\mathbf{Y}\|_F^2.\\
	\end{aligned}
\end{equation}

(2) DSTL-T dose not construct two silm tensor and utilizes traditional nuclear norm to replace tensor nuclear-norm regularization:
\begin{equation}\label{AT}
	\begin{aligned}
    \min_{\Omega} &\sum_{v=1}^{m}\|\mathbf{X}^{v}-\mathbf{W}^{v}(\mathbf{S}^{v}+\mathbf{H}^{v})\|_F^2+ \lambda_1\|\mathbf{S}^{v}\|_{1} \\
    &+\lambda_2\|\mathbf{H}^{v}\|_{\ast}+\lambda_3\|\mathbf{H}^{v}-\mathbf{C}^{v}\mathbf{Y}\|_F^2.\\
	\end{aligned}
\end{equation}

(3) DSTL-Y omits the consensus alignment indicator and does not use it to guide the disentanglement process:
\begin{equation}\label{AI}
	\begin{aligned}
    \min_{\Omega} &\sum_{v=1}^{m}\|\mathbf{X}^{v}-\mathbf{W}^{v}(\mathbf{S}^{v}+\mathbf{H}^{v})\|_F^2+ \lambda_1\|\mathcal{S}\|_{1} +\lambda_2\|\mathcal{H}\|_{\circledast}.\\
	\end{aligned}
\end{equation}
Similar constraints are imposed on the three variants as those in our DSTL. Fig.~\ref*{visualization} shows the t-SNE visualizations~\cite{van2008visualizing} of learned consensus alignment indicator embedding $\mathbf{Y}$ obtained by DSTL-S and DSTL on BBCSport, HW, Scene15, and Animal datasets. It can be observed that when using DSTL-S to obtain $\mathbf{Y}$, there are some clusters entangled with each other in the four datasets. However, we can see our DSTL is more robust and can achieve excellent performance in BBCSport and HW datasets when the negative influence of disentangled semantic-unrelated information is considered. 

Fig.~\ref{abl} shows the clustering performance of the three variants and our DSTL on nine datasets. We can observe that the performance of DSTL-S is not desirable on all datasets, eapecially on MITIndoor, CCV, and Animal. We know that there are considerable semantic-unrelated information in these datasets, which adversely affects the clustering performance. These results prove the effectiveness of the disentangled semantic-unrelated features learning once again. When dropping the tensor-based regularization, the clustering results of DSTL-T are notably inferior to DSTL, which demonstrates the effctiveness of utilizing tensor learning to explore high-order correlations. In NUSWIDEOBJ, it is evident that DSTL can achieve competitive results compared to DSTL-Y, confirming the effectiveness of learning consensus semantic alignment indicator to help align semantic-related representation across views, then guiding the process of feature disentanglement.

\section{Conclusion}
This paper proposes a novel fast tensor-based MVC method named disentangled slim tensor learning (DSTL), which integrates multi-view features disentanglement and slim tensor learning into a unified framework. To mitigate the negative influence of feature redundancy, DSTL disentangles the features of each view to learn both semantic-unrelated and semantic-related representations. It then constructs two slim tensors to alleviate the negative impact of semantic-unrelated information while also capturing the high-order consistency of different views. Additionally, the semantic-related representations are aligned across views to guide the disentanglement process by learning a consensus semantic alignment indicator. DSTL effectively reduce the space and time complexity compared with traditional tensor-based methods. Experimental results demonstrate the effectiveness and efficiency of our approach. One limitation of our method is that DSTL may struggle with incomplete multi-view data due to missing samples in practice. To address this, we will consider incorporating some recovery and completion learning strategies \cite{zhang2022low,zhang2023robust} in future work.


\ifCLASSOPTIONcaptionsoff
  \newpage
\fi

\bibliographystyle{IEEEtran}
\bibliography{TMM}

\begin{IEEEbiography}
	[{\includegraphics[width=1in,height=1.25in,clip,keepaspectratio]{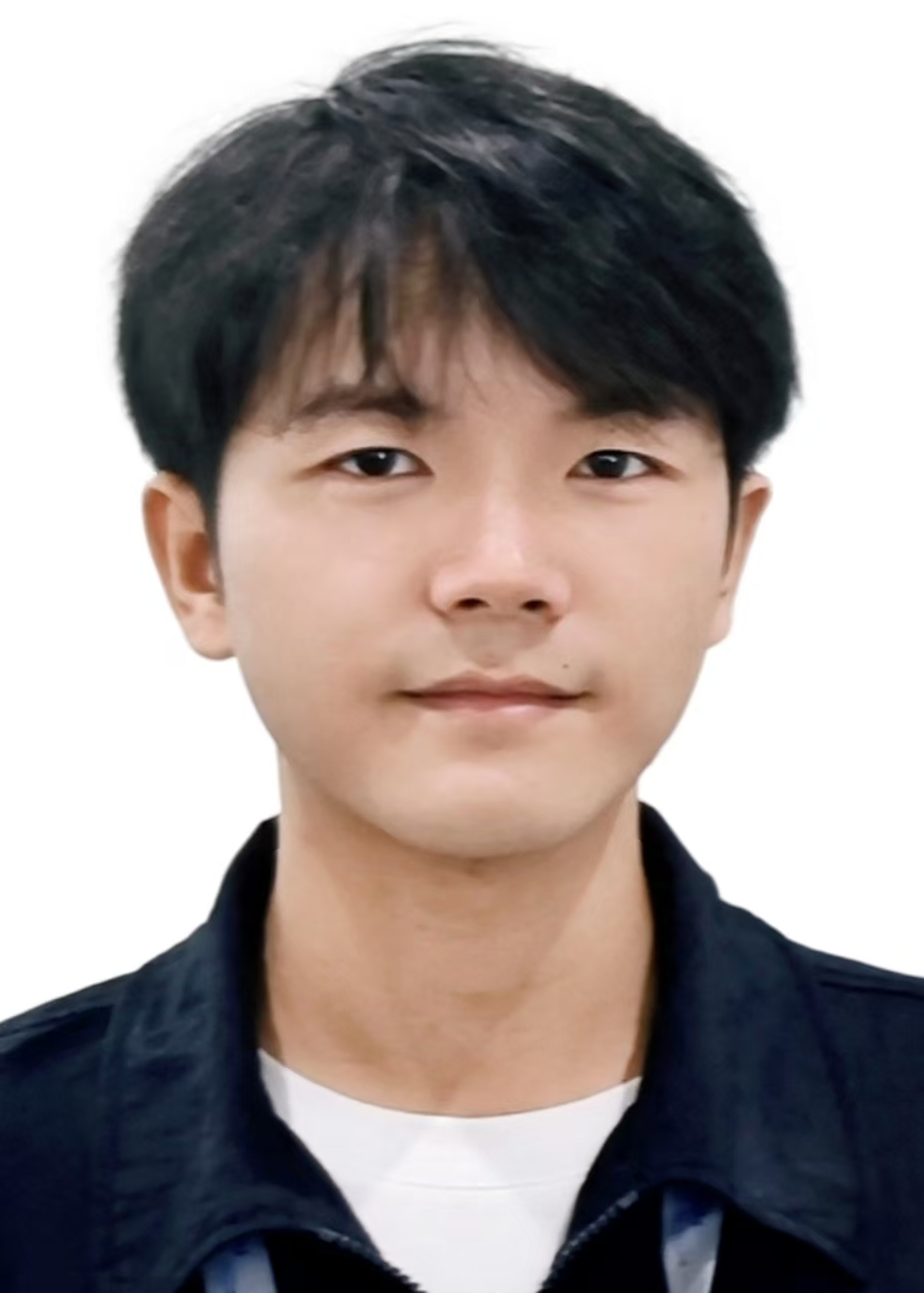}}]{Deng Xu}
    received the B.E. degrees in control science and engineering from Nanjing University, Nanjing, China, in 2023, where he is currently pursuing the Ph.D. degree with the Department of Control Science and Intelligence Engineering. His research interests include machine learning and data mining.
\end{IEEEbiography}

\begin{IEEEbiography}
	[{\includegraphics[width=1in,height=1.25in,clip,keepaspectratio]{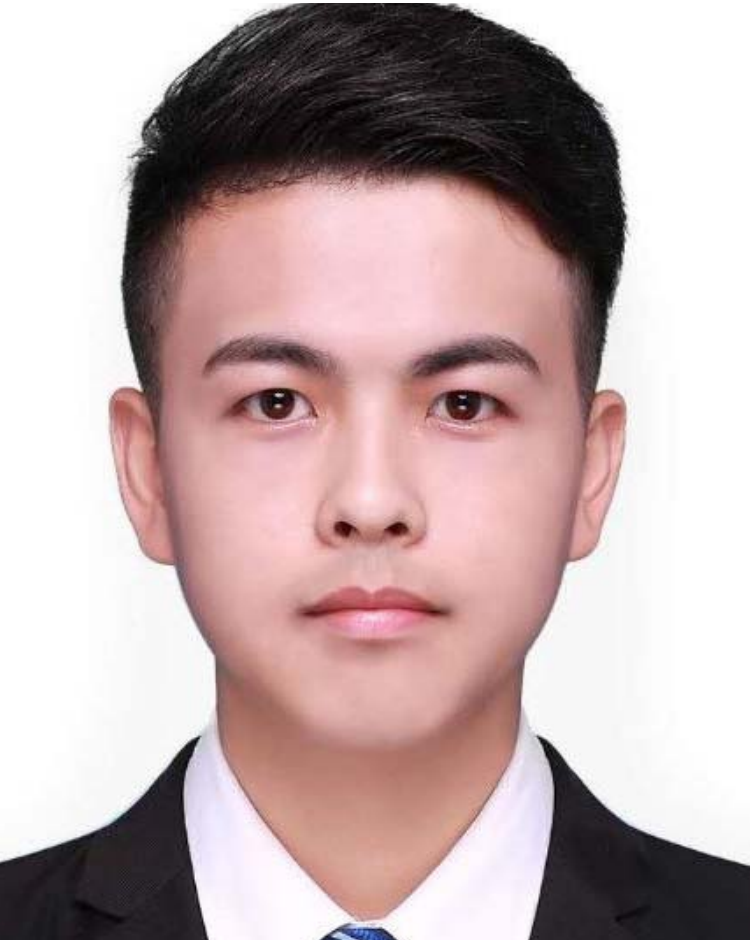}}]{Chao Zhang}
	received the B.E. and M.E. degrees in control science and engineering from Nanjing University, Nanjing, China, in 2018 and
	2021, respectively, where he is currently pursuing the Ph.D. degree with the Department of Control Science and Intelligence Engineering. His research interests include machine learning and pattern recognition.
\end{IEEEbiography}

\begin{IEEEbiography}
	[{\includegraphics[width=1in,height=1.25in,clip,keepaspectratio]{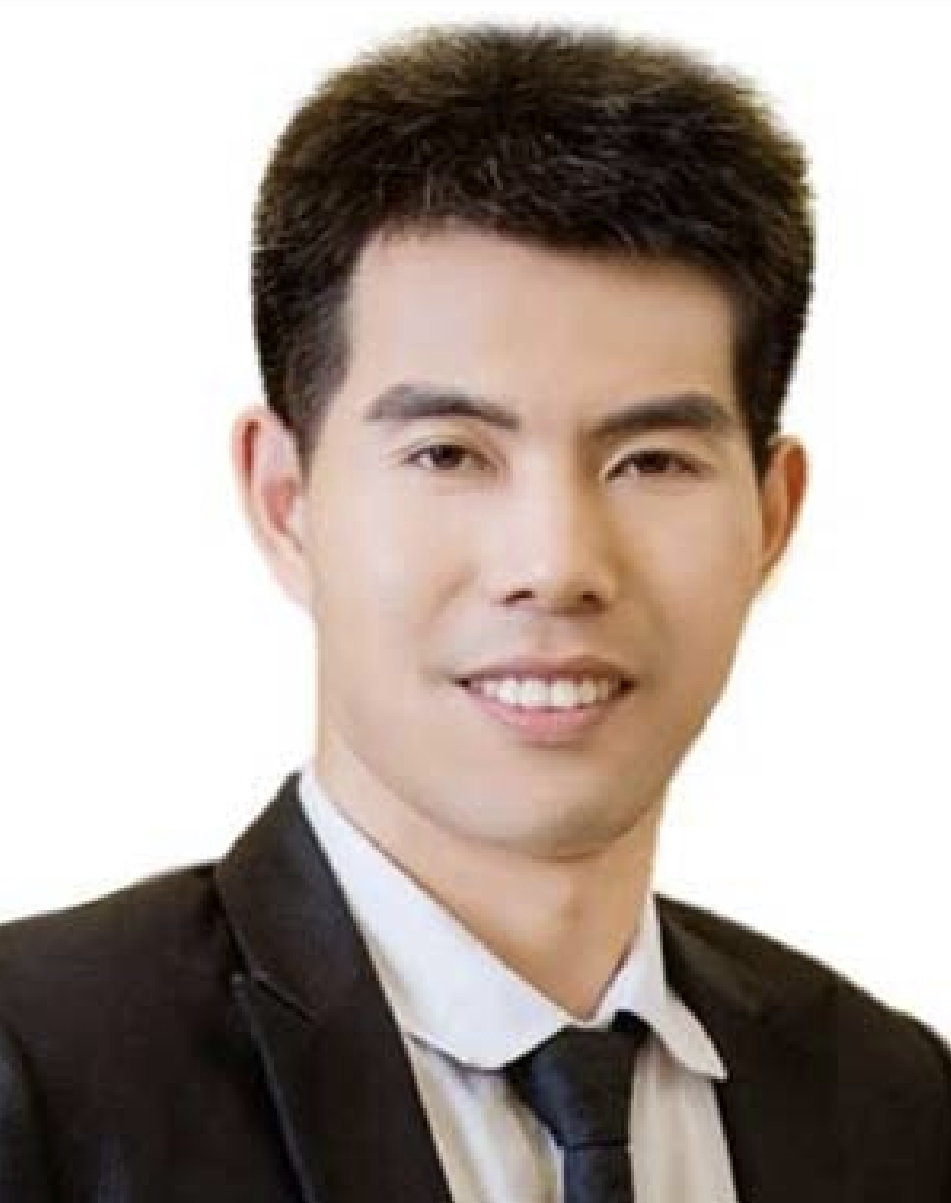}}]{Zechao Li}
	received the B.E. degree from the University of Science and Technology of China, in 2008 and the Ph.D. degree from the National Laboratory of Pattern Recognition, Institute of Automation, Chinese Academy of Sciences, in 2013. He is currently a professor with the Nanjing University of Science and Technology. His research interests include large-scale multimedia analysis, computer vision, Pattern Recognition, etc. He has authored over 70 papers in top-tier journals and conferences. He was a recipient of the best paper award in ACM
	Multimedia Asia 2020, and the best student paper award in ICIMCS 2018. He serves as the Associate Editor of IEEE TNNLS, Information Sciences, etc.
\end{IEEEbiography}
\begin{IEEEbiography}[{\includegraphics[width=1.0in,height=1.25in,clip,keepaspectratio]{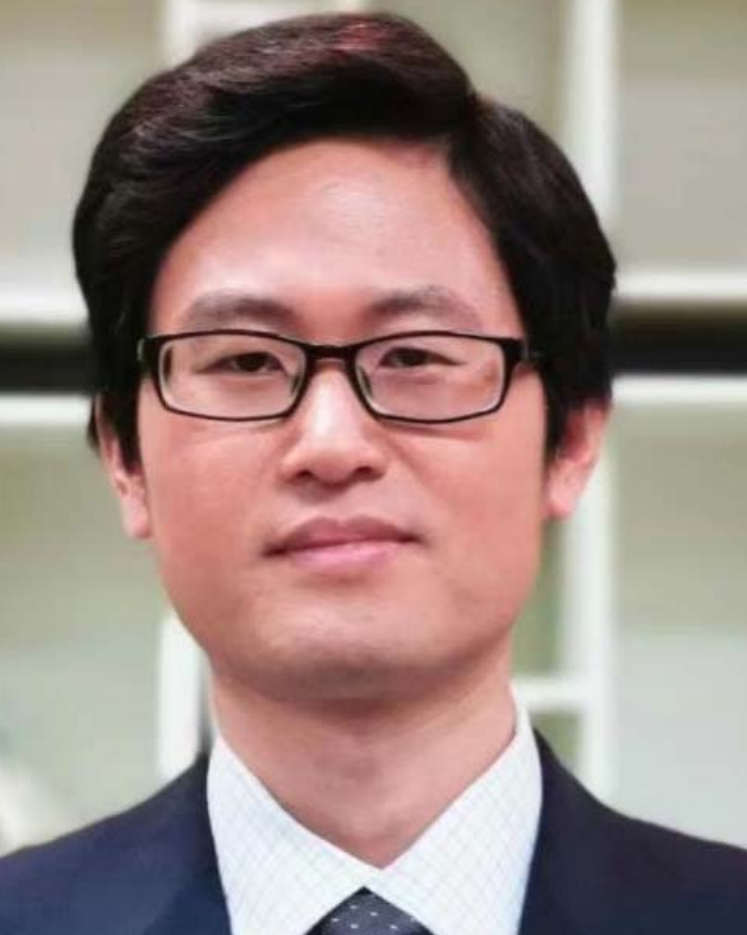}}]{Chunlin Chen}
	 received the B.E. degree in automatic control and Ph.D. degree in control science and engineering from the University of Science and Technology of China, Hefei, China, in 2001 and 2006, respectively. He is currently a professor and the vice dean of School of Management and Engineering, Nanjing University, Nanjing, China. He was a visiting scholar at Princeton University, Princeton, USA, from 2012 to 2013. He had visiting positions at the University of New South Wales, Canberra, Australia, and the City University of Hong Kong, Hong Kong, China. His recent research interests include machine learning, pattern recognition, intelligent information processing, and quantum control. He is the Chair of Technical Committee on Quantum Cybernetics, IEEE Systems, Man and Cybernetics Society. 
\end{IEEEbiography}

\begin{IEEEbiography}
	[{\includegraphics[width=1in,height=1.25in,clip,keepaspectratio]{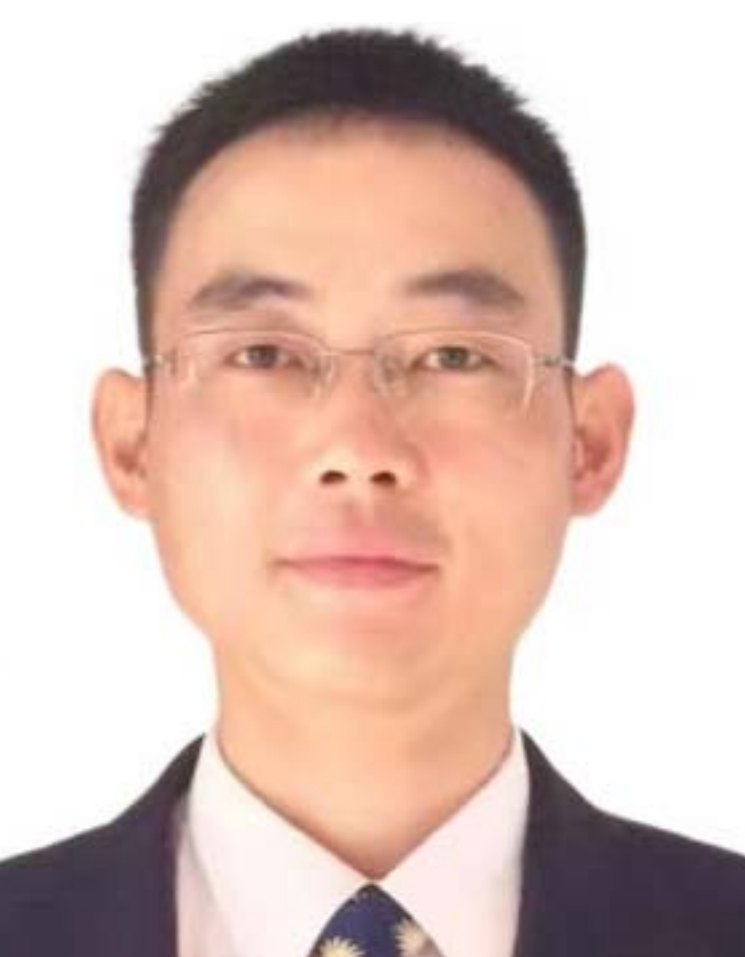}}]{Huaxiong Li} received the M.E. degree in control theory and control engineering from Southeast University, Nanjing, China, in 2006, and Ph.D. degree from Nanjing University, Nanjing, China, in 2009. He is currently a Professor with the Department of Control Science and Intelligence Engineering, Nanjing University, Nanjing, China. He was a visiting scholar at the Department of Computer Science, University of Regina, Canada, from 2007 to 2008, and a visiting scholar at the University of Hong Kong, Hong Kong, China, in 2010. His current research interests include machine learning, data mining, and computer vision. He has published more than 100 peer-reviewed papers in IEEE TKDE, IEEE TPAMI, IEEE TNNLS, etc. He is a Committee Member of China Association of Artificial Intelligence (CAAI) Machine Learning Committee / Granular Computing and Knowledge Discovery (GCKD) Committee, a CAAI Senior Member, a China Computer Federation (CCF) Distinguished Member, and a Committee Member of JiangSu association of Artificial Intelligence (JSAI) Pattern Recognition Committee.
\end{IEEEbiography}

\end{document}